%% file: main.tex
\documentclass[10pt,twocolumn,letterpaper]{article}

\usepackage{wacv}              


\definecolor{wacvblue}{rgb}{0.21,0.49,0.74}
\usepackage[pagebackref,breaklinks,colorlinks,allcolors=wacvblue]{hyperref}

\def\confName{WACV}
\def\confYear{2027}

\usepackage{url}
\usepackage{multirow} 
\usepackage{graphicx}
\usepackage{subcaption}
\usepackage{algorithm}

\usepackage{algorithmic}
\usepackage{adjustbox}

\usepackage{tikz}
\usepackage{amsmath}
\usepackage{mathtools}
\usepackage{amsthm}
\usepackage{amssymb} 
\usepackage{xcolor} 
\usepackage[table]{xcolor}
\usepackage{url}
\usepackage{booktabs}
\usepackage{tabularx}
\usepackage{amsmath}
\usepackage{listings}
\usepackage{mathrsfs}
\usepackage{arydshln}
\usepackage{bm}
\usepackage{pifont}
\usepackage[most]{tcolorbox}
\newtcolorbox{takeaway}{
  colback=blue!3,     
  colframe=blue!70!black, 
  arc=2mm,            
  boxrule=0.8pt       
}
\newtheorem{definition}{Definition}
\newtheorem{theorem}{Theorem}[section]
\newtheorem{proposition}[theorem]{Proposition}

\usepackage{cancel}

\definecolor{beige}{RGB}{250,240,210}
\definecolor{peach}{RGB}{255,220,180}
\definecolor{deeppeach}{RGB}{180, 85, 45}
\definecolor{softred}{RGB}{220, 110, 110}
\definecolor{lightpink}{RGB}{255, 220, 225}
\definecolor{mint}{RGB}{200,250,230}
\definecolor{lightpurple}{RGB}{235,210,255} 
\definecolor{purple}{RGB}{145,115,180}
\definecolor{softblue}{RGB}{0, 160, 220}
\definecolor{deepblue}{RGB}{70, 130, 180}
\definecolor{softgreen}{RGB}{90,150,125}
\definecolor{deepgreen}{RGB}{0,128,0} 
\definecolor{midgray}{gray}{0.8}
\definecolor{lightgray}{gray}{0.87}
\definecolor{darkred}{RGB}{192, 0, 0}
\newcommand{\cosine}{Similarity}
\newcommand{\mi}{\textbf{MI-Pruner}}
\newcommand{\mia}{\textbf{{MI-Attention}}}

\usepackage[normalem]{ulem}
\newcommand{\sota}[1]{\textbf{{#1}}}
\newcommand{\subsota}[1]{{\uline{#1}}}
\newcommand{\vvv}[1]{\textcolor{gray}{\ensuremath{_\text{#1}}}}

\title{\mi: Crossmodal Mutual Information-guided Token Pruner \\for Efficient MLLMs}

\author{
Jiameng Li$^1$~~~~Aleksei Tiulpin$^2$~~~~Matthew B. Blaschko$^1$\\
$^1$KU Leuven, ~~$^2$Weill Cornell Medicine
\\
\url{https://github.com/renaissanceee/MI-Pruner}
}

\begin{document}
\maketitle
\input{sec/0_abstract}

\input{sec/1_intro}

\input{sec/2_method}

{
    \small
    \bibliographystyle{ieeenat_fullname}
    \bibliography{main}
}
\input{sec/3_appendix}

\end{document}

%% file: sec/0_abstract.tex
\begin{abstract}

The redundancy of visual tokens in multimodal large language models (MLLMs) motivates training-free token pruning techniques for efficient inference.
Existing methods typically rank visual tokens by attention salience and retain only the most salient ones. However, salience-based pruning is susceptible to sink tokens and positional bias, overlooking the diversity of the selected set. Moreover, extracting attention maps incurs additional overhead. 
In this paper, we propose \mi, which derives Pointwise Mutual Information (PMI) based relevance scores in the projection space to reflect crossmodal relevance and intra-modal redundancy.
Grounded in information theory, our approach identifies the most informative tokens by explicitly measuring feature-level dependency.
Without reliance on internal attention maps or architectural modifications, \mi~can be seamlessly applied to any pretrained MLLMs for inference acceleration.
Extensive experiments on LLaVA1.5 and Qwen-series demonstrate that our approach achieves a favorable performance-efficiency trade-off. Specifically, we preserve ${>}97\%$ performance on LLaVA1.5-7B at a pruning rate of 88.9\%.

\end{abstract}

%% file: sec/1_intro.tex
\section{Introduction}
\label{sec:intro}

Modern multimodal models have rapidly grown in scales and capabilities \citep{Qwen3VL,chen2024internvl,liu2024llavanext}, leading to substantial computational costs. 
Notably, the quadratic complexity of the attention mechanism makes inference increasingly expensive as the number of tokens grows when processing high-resolution visual inputs \citep{li2024llava}. To address it, token pruning has emerged as a critical technique for efficient inference, aiming to retain only the most informative tokens while discarding redundant ones \citep{ye2025atp,cao2024madtp}. 
By reducing sequence length in the forward pass, the pruning not only alleviates computation and memory consumption, but also enables scalable deployment of large multimodal models in latency- and resource-constrained scenarios. However, it remains a challenge to identify informative tokens \citep{fan2025visipruner}. Erroneously pruning evidence tokens leads to LLM hallucination, where responses are generated merely by prior knowledge in LLMs rather than input evidence \citep{bai2024hallucination}.

\begin{figure*}[t]
\centering
\includegraphics[width=\linewidth]{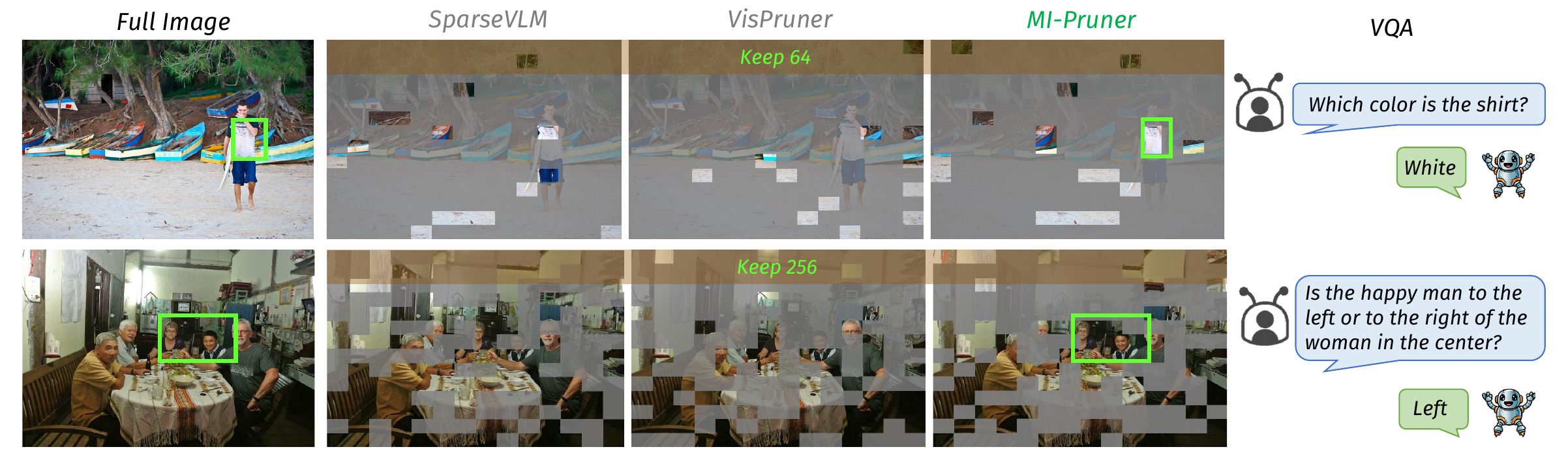}
    \vspace{-0.4cm}
\caption{\textbf{Pruning visualization on LLaVA1.5-7B with different budgets.} MI-Pruner consistently identifies and preserves the queried regions, whereas other methods partially miss relevant information. 
}
\label{fig:vis_llava}
\end{figure*}

Motivated by studies on multimodal information flow \citep{zhang2025cross,zhang2025shallow}, salience-based pruning emerges by collecting attention maps from vision encoders or LLMs.
Some work \citep{chen2024image,zhang2024sparsevlm,liprim} considers visual dilution after LLM shallow layers and selects tokens by analyzing the cross-attention matrix in LLM decoders. 
Yet, the entangled multimodal semantics in LLMs contain inherent noise \citep{zhang2025beyond,cai2024flashvlm} to distort the importance score.
As a result, recent approaches \citep{zhang2025beyond,yang2025visionzip,cai2024flashvlm} turn to visual cues for early pruning, \textit{i.e.}, self-attention scores extracted in the vision encoder. 
From the perspective of information theory, the attention mechanism \citep{vaswani2017attention} employs learnable QKV matrices to capture token-wise dependencies, and the attention scores work as an approximation to Pointwise Mutual Information (PMI). Eventually, the attention module learns an implicit distribution that maximizes the Mutual Information between queries and values. 
Despite the progress, there exist several shortcomings in attention-based pruning.
\textbf{\textit{(i) Heuristic metrics with miscalibrated attention:} }
The raw attention scores don't fully equal semantic importance due to the positional bias \citep{wen2025token}, attention sinks \citep{xiao2023efficient,yangunderstanding} and multi-head diversity \citep{zhang2025shallow,kang2025see}. 
As observed, attention maps have outliers incongruous with local semantics \citep{gu2025attention}, and the multi-head average obliterates the head-specific roles \citep{yangunderstanding,kang2025see}. 
Therefore, a necessary calibration step is missing in previous salience-based methods \citep{darcet2023vision,laurenccon2024matters,yanvision}.
\textbf{\textit{(ii) Imbalanced diversity:}} 
Prioritizing high attention scores tends to preserve salient but
overly similar tokens \citep{zou2025don}. The imbalanced concern of salience and diversity leads to a substantial performance drop under high pruning ratios, since the visual importance is not an isolated metric but depends on textual queries and visual contexts.
\textbf{\textit{(iii) Implementation bottlenecks:}}
The inner-LLM operations \citep{chen2024image,zhang2024sparsevlm} are incompatible
with the latest attention cores like FlashAttention \citep{dao2023flashattention2}, while collecting attention maps from visual encoders increases memory overhead and overlooks the multimodal nature.

Unlike the accumulated salience from attention maps \citep{yang2025visionzip,zhang2025beyond}, we propose a surgical method derived from Mutual Information and conduct greedy search using our MI-based relevance.
In the projection space, we construct temperature-scaled similarity matrices from multimodal features, then apply softmax to estimate conditional and marginal probabilities. These probabilities are subsequently used to compute Pointwise Mutual Information and aggregated for our MI-based relevance scores. Finally, we conduct greedy selection by crossmodal and intra-modal MI-based relevance scores.
Different from the attention mechanism, we derive probability scores and token-wise dependency in a training-free and surgical manner.
As a result, our \mi~circumvents extra learnable QKV matrices and the collection of attention maps from vision encoders \citep{yang2025visionzip} or LLMs \citep{chen2024image}. 
Grounded in information quantities \citep{iyer2021submodular,spadaro2023shannon,gao2020rethinking,potkins2024improve}, our MI-based scoring function effectively identifies irreplaceable and query-relevant vision tokens. 
As a model-agnostic token reduction approach, \mi~supports non-intrusive pruning and can be easily applied to a wide range of MLLMs following the Enc-MLP-Dec paradigm. 
Experiments across various benchmarks demonstrate that our method achieves highly competitive performance 
and significantly accelerates inference.
Moreover, we present \mia, which integrates our MI-based scoring and attention salience for flexible usage.

In summary, our \textbf{contributions} are:
\begin{itemize}
    \item We propose \mi, an attention-free visual token pruning method, which brings inference speedup and captures visual cues aligned with the textual query.
    \item As a model-agnostic module, \mi~can be seamlessly integrated into various off-the-shelf MLLMs with controllable budgets. 
    \item Empirical results on LLaVA1.5 and Qwen-series indicate our effectiveness and efficiency, which proves that attention is not the only measure of semantic importance. 
\end{itemize}

\section{Related Work}
\label{sec:related}

\paragraph{Salience-based pruning.}
Many works \citep{chen2024image,fan2025visipruner,lin2025boosting} observed that image tokens receive less attention in LLMs' deep layers, which boosts attention salience-based visual compression in LLMs.
As a pioneer, FastV \citep{chen2024image} discards non-salient [\texttt{vis}] tokens guided by cross-attention scores in LLMs. The following work, SparseVLM \citep{zhang2024sparsevlm} selects a few [\texttt{text}] tokens as raters, then calculates the rank of attention matrices for dynamically pruning in LLMs.
Another branch argues that purely visual cues are sufficient to indicate the informative patches. VisionZip \citep{yang2025visionzip} first selects image tokens with high [\texttt{CLS}] attention in the last layer of the vision encoder (round-1) and then merges the remaining tokens by semantic similarity (round-2) to compress the LLM inputs. Similarly, VisPruner \citep{zhang2025beyond} progressively removes duplicates by similarity in round-2.
Despite working in different stages, 
attention salience-based methods tend to suffer from semantic miscalibration, diversity imbalance and implementation bottlenecks.

\begin{figure*}[h]
\centering
\includegraphics[width=0.99\linewidth]{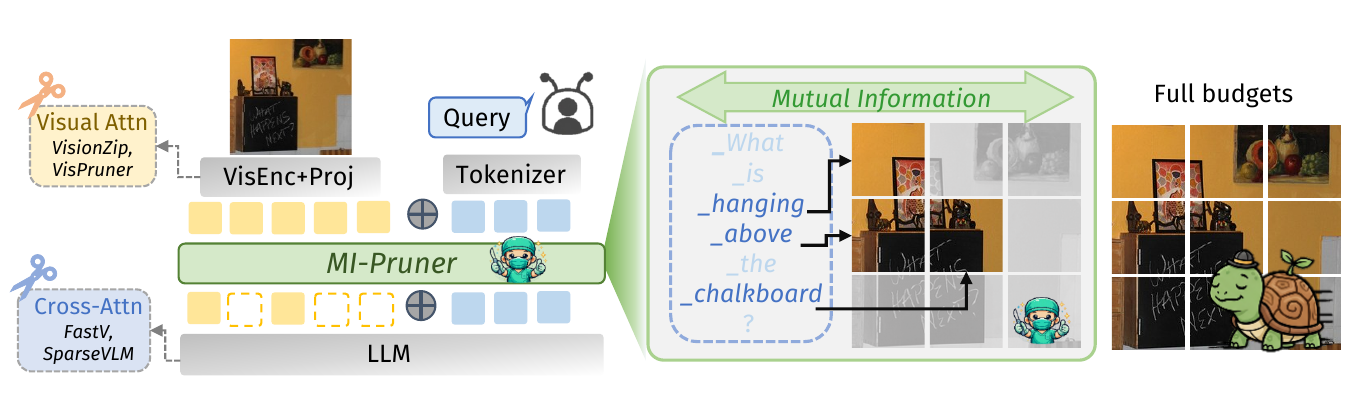}
\caption{
\textbf{Overview.} Previous methods prune tokens by collecting attention maps from vision encoders or LLM decoders. Motivated by Mutual Information, MI-Pruner estimates cross- and intra-modal dependency in projection space, achieving a desired performance-efficiency trade-off.
}
\label{fig:pipeline}
\end{figure*}

\paragraph{Subset-based pruning.} 
A bunch of work stops chasing the highlight tokens from attention maps but turns to study the problem of subset coverage, \textit{e.g.}, similarity or conditional diversity measures.
DART \citep{wen2025stop} introduces pivot tokens in LLMs to emphasize the (dis-)similarity over importance, while ApET \citep{ma2026apet} 
turns to measure how well the selected set covers a new token. 
Divprune \citep{alvar2025divprune} and CDPruner \citep{zhang2025cdpruner} leverage Min-Max diversity and conditional similarity to score visual tokens, and HoloV \citep{zou2025don} considers both attention and diversity for a balanced metric.
Akin to ours, MMToK \citep{dong2025mmtok} constructs an energy-based similarity matrix, yet their analysis ends at empirical coverage. 
As a classic information processing tool, Mutual Information was also introduced into token pruning.
For instance, AutoPrune \citep{wang2025autoprune} interprets attention scores as probabilities in their MI scores, assuming uniform text tokens.
Under conditional Gaussian distributions, TrimTokenator \citep{zhang2025trimtokenator} adopts the L2-norm as MI proxy for visual pruning.
In comparison, \mi~conducts softmax over projection-space features to derive probabilities used in MI-based relevance scores,
which avoids latency from collecting attention maps and the miscalibration issue \citep{huang2025ppe}. 
We do not claim to be the first MI-based token pruning method. Instead, we introduce an efficient PMI-based token compression framework that is attention-free, computationally efficient, and readily applicable to a wide range of MLLMs.

\section{Preliminaries}
\label{sec:pre_mllm}

\paragraph{MLLM architectures.}
The general MLLM architectures \citep{liu2024improved,Qwen3VL} follow an Enc-MLP-Dec paradigm, including a vision encoder $\mathbf{E_v}$ with an MLP projection module $g$, a tokenizer $\mathbf{E_T}$, and an LLM decoder. Given an image $I$ and a prompt $P$ (system instructions omitted), multimodal features $\mathcal{V, T}$ are extracted as:
\begin{align}
\mathcal{V} &:= \{\mathbf{v}_i\}_{i=1}^{N_V} = g(\mathbf{E_{V}}(I)) \in \mathbb{R}^{N_V \times d}, 
\quad \mathbf{v}_i \in \mathbb{R}^{d}, \\
\mathcal{T} &:= \{\mathbf{t}_j\}_{j=1}^{N_T} = \mathbf{E_{T}}(P) \in \mathbb{R}^{N_T \times d}, 
\quad \mathbf{t}_j \in \mathbb{R}^{d}.
\end{align}
Then, $N_V$ image features and $N_T$ text features are concatenated as inputs for LLM decoding:
\begin{equation}
\mathcal{X}:= \mathcal{V}\oplus \mathcal{T} \in \mathbb{R}^{(N_V+N_T) \times d}.
\label{eq:concat}
\end{equation}

\paragraph{Visual attention for pruning.}
To assess token importance, methods based on visual cues \citep{zhang2025beyond,yang2025visionzip} typically employ a two-step pruning strategy: (i) aggregate attention scores from [\texttt{CLS}] or other image tokens from $\mathcal{V}$ as $\mathrm{Attn}_{i}$, to determine the initial pruning budget; (ii) recycle from remaining tokens $\mathcal{V}_\mathrm{re}$ by their similarity $S_i$.
Let $\text{cls}$ denote the [\texttt{CLS}] index in the attention matrix $\mathbf{A} \in \mathbb{R}^{N_V \times N_V}$, and let $\tilde{\mathbf{v}}_{i} \in \mathbb{R}^d$ denote the normalized visual features before projection.
The two-step importance metrics are formulated as:
\begin{align}
\mathrm{Attn}_{i} &= \mathbf{A}_{\text{cls},i}~\text{or}~ \sum_{j\ne i} \mathbf{A}_{ji},
\label{eq:a_i}\\
S_i &= \sum_{j\ne i, \tilde{\mathbf{v}}_j \in \mathcal{V}_\mathrm{re}}(\tilde{\mathbf{v}}_i^\top \tilde{\mathbf{v}}_j),
\label{eq:s_i}
\end{align}
where $\cdot^\top$ denotes the transpose operator.
By replacing Eqn.~(\ref{eq:s_i}) with our MI-based scoring function, our approach can be integrated into attention salience-based methods like VisPruner \citep{zhang2025beyond} and VisionZip \citep{yang2025visionzip}. We denote this variant as \mia~ in the following.

%% file: sec/2_method.tex
\section{Methods}
\label{sec:methods}

\subsection{Overview}
\label{subsec:methods_sim}

To preserve query-relevant tokens while pruning visually redundant ones, we introduce Mutual Information-based scoring functions to model cross- and intra-modal relevance. Our pruning occurs in the projection space, where visual and textual features are aligned but have not yet undergone multimodal interaction.
As shown in Fig.~\ref{fig:pipeline}, our pruning starts from visual features
$\mathcal{V} = \{\mathbf{v}_{i}\}_{i=1}^{N_V}$,
and textual features
$\mathcal{T} = \{\mathbf{t}_j\}_{j=1}^{N_T}$, where $\mathbf{v}_i,\mathbf{t}_j \in \mathbb{R}^{d}$.
The objective is to prune visual patches that are weakly correlated
with the textual semantics, yet highly redundant among selected patches. 
To save space, we adopt a unified notation: $\mathrm{s} \in \{\text{cross}, \text{self}\}, \mathbf{x} \in \{\mathbf{t}, \mathbf{v}\}, N \in \{N_T, N_V\}$.
To make the objective tractable, we adopt two simplifying assumptions: (i) conditional independence, (ii) uniform prior probability. These assumptions are primarily introduced to improve computational efficiency and enhance generalization, although they may not strictly hold in practice.
Detailed discussions can be found in App.~\ref{subsec:app_theory_cond_indep} to \ref{subsec:app_theory_prior}.

Our pruning consists of 4 steps. Firstly, we construct similarity matrices from normalized projection-space features. Then, we derive conditional probabilities $p({\mathbf{x}}_j \mid {\mathbf{v}}_i)$ by applying a softmax operation over the similarity matrix, and formulate marginal probabilities $p({\mathbf{t}}_j)$ by assuming a uniform prior as \cite{wang2025autoprune}. The third step is to calculate Pointwise Mutual Information from the derived probabilities. Finally, the PMI measures are aggregated into MI-based relevance scores used for greedy selection. 
See Algorithm~\ref{alg:mi_pruner} for our pipeline.

\begin{figure}[h]
  \centering 
\includegraphics[width=\columnwidth]{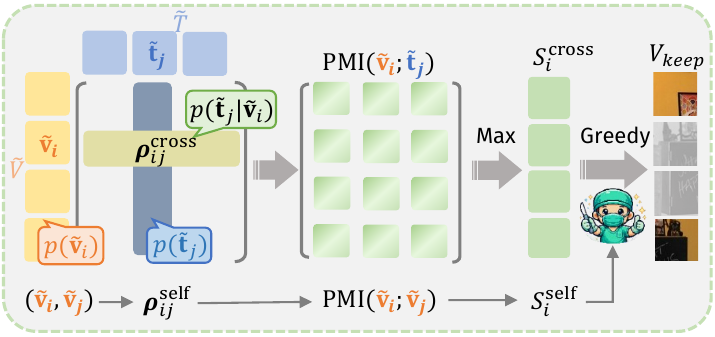} 
\vspace{-0.3cm}
  \caption{ 
    \textbf{Our MI-based scoring function.}
    All [\texttt{vis}] tokens are flattened for illustration. 
    We incorporate crossmodal (top) and intra-modal (bottom) PMI scores as the visual importance metric.
  }
  \label{fig:pipeline_mi}
  \vspace{-0.3cm} 
\end{figure}

\subsection{\mi}
\label{subsec:methods_mi}

\paragraph{Similarity matrices.} 
We first conduct normalization to $\mathcal{V}$ and $\mathcal{T}$, yielding normalized features $\tilde{\mathcal{V}}$ and $\tilde{\mathcal{T}}$. 
Based on Boltzmann distributions, the similarity matrices for vision-text ($\bm{\rho}_{ij}^\mathrm{cross}$) and vision-vision ($\bm{\rho}_{ij}^\mathrm{self}$) pairs are computed as:
\begin{align}
\bm{\rho}_{ij}^\mathrm{cross} = \frac{\tilde{\mathbf{v}}_i^\top \tilde{\mathbf{t}}_j}{\tau}, 
\quad \bm{\rho}_{ij}^\mathrm{self} = \frac{\tilde{\mathbf{v}}_i^\top \tilde{\mathbf{v}}_j}{\tau},
\label{eq:sim}
\end{align}
where the temperature $\tau$ controls the sharpness of the distribution. For simplicity, we adopt a shared $\tau$ for both matrices, though it's not mandatory.
Treating these similarity scores 
$\{\bm{\rho}_{ij}^\mathrm{s}\}_{j=1}^{N}$ 
as logits, we define the energy-based crossmodal and internal conditional distribution.

\paragraph{Conditional distribution.}
Applying a softmax operation along the second dimension, we obtain conditional probabilities $p(\tilde{\mathbf{t}}_j \mid \tilde{\mathbf{v}}_i),~p(\tilde{\mathbf{v}}_j \mid \tilde{\mathbf{v}}_i)$ in a unified formula: 
\begin{align}
p(\tilde{\mathbf{x}}_j \mid \tilde{\mathbf{v}}_i) =\operatorname{softmax}_j(\bm{\rho}_{ij}^\mathrm{s})= \frac{\exp(\bm{\rho}_{ij}^\mathrm{s})}{\sum_{k=1}^{N} \exp(\bm{\rho}_{ik}^\mathrm{s})}.
\label{eq:softmax}
\end{align}
It quantifies the semantic alignment between $(\tilde{\mathbf{v}}_i, \tilde{\mathbf{t}}_j)$ and the internal diversity among $(\tilde{\mathbf{v}}_i, \tilde{\mathbf{v}}_j)$,
as illustrated by the row operation in Fig.~\ref{fig:pipeline_mi}.

\paragraph{Marginal distribution.} 
Following \cite{wang2025autoprune}, we assume a uniform prior $p(\tilde{\mathbf{v}}_j){=}\frac{1}{N_V}$, where visual embeddings occur with equal probabilities to avoid spatial bias.
According to the law of total probability, the marginal distribution of text embeddings is expressed as:
\begin{equation}
p(\tilde{\mathbf{t}}_j) = \frac{1}{N_V} \sum_{i=1}^{N_V} p(\tilde{\mathbf{t}}_j \mid \tilde{\mathbf{v}}_i),
\label{eq:margin_t}
\end{equation}
as illustrated by the column operation in Fig.~\ref{fig:pipeline_mi}. More details can be found in App.~\ref{subsec:app_theory_prior} and ~\ref{subsec:app_exp_more}.

\paragraph{Pointwise Mutual Information.}
Pointwise Mutual Information measures the token-wise relevance. We adopt a unified notation $\tilde{\mathbf{x}}_j\in\{\tilde{\mathbf{t}}_j,\tilde{\mathbf{v}}_j\}$ to denote pairwise PMI:
\begin{equation}
\text{PMI}(\tilde{\mathbf{v}}_i; \tilde{\mathbf{x}}_j) = \log \frac{p(\tilde{\mathbf{x}}_j \mid \tilde{\mathbf{v}}_i)}{p(\tilde{\mathbf{x}}_j)}.
\end{equation}
To elaborate on that, $\text{PMI}(\tilde{\mathbf{v}}_i; \tilde{\mathbf{t}}_j)$ quantifies the crossmodal semantic alignment, while $\text{PMI}(\tilde{\mathbf{v}}_i; \tilde{\mathbf{v}}_j)$ captures the intra-modal redundancy. The pairwise PMI scores are aggregated for our MI-based relevance.

\paragraph{Mutual Information-based relevance.}
Eqn.~(\ref{eq:mi_2}) transfers pairwise PMI into MI by a weighted average, and justifies the following greedy search.
In token pruning, the weighted average works as \textit{global aggregation} over all text tokens or selected image tokens, which brings the consensus MI-based relevance $S_i^\mathrm{s}$~($\mathrm{s} \in \{\text{cross}, \text{self}\}$):
\begin{align}
S_i^\mathrm{s}
&= 
\sum_{j=1}^{N_X}\underbrace{p(\tilde{\mathbf{v}}_i; \tilde{\mathbf{x}}_j)}_{p(\tilde{\mathbf{x}}_j \mid \tilde{\mathbf{v}}_i)} \cdot
{\mathrm{PMI}(\tilde{\mathbf{v}}_i; \tilde{\mathbf{x}}_j)}.
\label{eq:proxy_avg_1}
\end{align}
Yet, the consensus over many text tokens or image patches brings smoothing effects with diluted significance. We instead implement a \textit{maximal aggregation} strategy pointing to keywords and discriminative patches.
\begin{align}
S_i^\mathrm{cross}
&= 
\max_{\tilde{\mathbf{t}}_j \in \tilde{\mathcal{T}}}
{\mathrm{PMI}(\tilde{\mathbf{v}}_i; \tilde{\mathbf{t}}_j)},
\label{eq:proxy_cross}
\\
S_i^\mathrm{self}&=
\max_{\tilde{\mathbf{v}}_j \in \tilde{\mathcal{V}}_\mathrm{S}}
{\mathrm{PMI}(\tilde{\mathbf{v}}_i; \tilde{\mathbf{v}}_j)}.
\label{eq:proxy_self}
\end{align}
And we define an MI-based scoring function $S_i$ as the marginal gain of adding $\mathbf{v}_i$ based on Eqn.~(\ref{eq:proxy_cross}-\ref{eq:proxy_self}):
\begin{align}
&S_i
= \lambda S_i^\mathrm{cross}{-}(1{-}\lambda) \cdot S_i^\mathrm{self}
\label{eq:score_1}
\\
&=\lambda\cdot
\underbrace{\max_{\tilde{\mathbf{t}}_j \in \tilde{\mathcal{T}}}
{\mathrm{PMI}(\tilde{\mathbf{v}}_i; \tilde{\mathbf{t}}_j)}}_{\text{relevance}\rightarrow\text{sorting}}{-}
(1{-}\lambda) 
\underbrace{\max_{\tilde{\mathbf{v}_j} \in \tilde{\mathcal{V}}_\mathrm{S}}
{\mathrm{PMI}(\tilde{\mathbf{v}}_i; \tilde{\mathbf{v}}_j)}}_{\text{redundancy}\rightarrow\text{greedy}}.
\label{eq:score_2}
\end{align}
\begin{algorithm}[h]
\caption{\mi}
\label{alg:mi_pruner}
\begin{algorithmic}[1]

\STATE \textbf{Input:}
Visual and textual embeddings $\mathcal{V}=\{\mathbf{v}_i\}_{i=1}^{N_V}$, $\mathcal{T}=\{\mathbf{t}_j\}_{j=1}^{N_T}$,
temperature $\tau$,
budget $N_{\mathrm{keep}}$

\STATE \textbf{Output:}
The selected visual subset $\mathcal{V}_{\mathrm{keep}}$
\STATE \textcolor{gray}{\textit{// Similarity matrices}}

\STATE $\tilde{\mathbf{v}}_i, \tilde{\mathbf{t}}_j  \leftarrow \mathbf{v}_i / \lVert \mathbf{v}_i \rVert, \mathbf{t}_j / \lVert \mathbf{t}_j \rVert$
\STATE $\boldsymbol{\rho}^\mathrm{cross}_{ij}  ,\boldsymbol{\rho}^\mathrm{self}_{ij} \leftarrow \tilde{\mathbf{v}}_i^\top \tilde{\mathbf{t}}_j / \tau, \tilde{\mathbf{v}}_i^\top \tilde{\mathbf{v}}_j / \tau$

\STATE \textcolor{gray}{\textit{// Derive distributions (softx means softmax)}}

\STATE {$p(\tilde{\mathbf{t}}_j | \tilde{\mathbf{v}}_i), p(\tilde{\mathbf{v}}_j | \tilde{\mathbf{v}}_i)  \leftarrow \text{softx}_j(\boldsymbol{\rho}^\mathrm{cross}_{ij}), \text{softx}_j(\boldsymbol{\rho}^\mathrm{self}_{ij})$}

\STATE $p(\tilde{\mathbf{v}}_j), p(\tilde{\mathbf{t}}_j) \leftarrow \frac{1}{N_V},\frac{1}{N_V}
\sum_{i=1}^{N_V} p(\tilde{\mathbf{t}}_j \mid \tilde{\mathbf{v}}_i)$

\STATE \textcolor{gray}{\textit{// Pointwise Mutual Information (PMI)}}

\STATE {$\mathrm{PMI}(\tilde{\mathbf{v}}_i;\tilde{\mathbf{t}}_j), \mathrm{PMI}(\tilde{\mathbf{v}}_i;\tilde{\mathbf{v}}_j)
\leftarrow \log
\frac{p(\tilde{\mathbf{t}}_j \mid \tilde{\mathbf{v}}_i)}
 {p(\tilde{\mathbf{t}}_j)},\log
\frac{p(\tilde{\mathbf{v}}_j \mid \tilde{\mathbf{v}}_i)}
     {p(\tilde{\mathbf{v}}_j)}$}

\STATE \textcolor{gray}{\textit{// Greedy selection}}
\STATE $\text{Initialize } \mathcal{V}_\mathrm{S} = \emptyset$
\WHILE{$|\mathcal{V}_{\mathrm{S}}| < N_{\mathrm{keep}}$}
    \FOR{each $\mathbf{v}_i \in \mathcal{V} \setminus \mathcal{V}_{\mathrm{S}}$}

        \STATE $S_i \leftarrow
        \lambda \max_{\tilde{\mathbf{t}}_j \in \tilde{\mathcal{T}}}
    {\mathrm{PMI}(\tilde{\mathbf{v}}_i; \tilde{\mathbf{t}}_j)}$
    \\ \hspace{2em} $-
    (1{-}\lambda) \max_{\tilde{\mathbf{v}}_j \in \tilde{\mathcal{V}}_\mathrm{S}}
        {\mathrm{PMI}(\tilde{\mathbf{v}}_i; \tilde{\mathbf{v}}_j)}$
    \ENDFOR
    \STATE $\mathbf{v}^\star \leftarrow \arg\max_{\mathbf{v}_i} S_i, \text{and}~\mathcal{V}_{\mathrm{S}} \leftarrow
    \mathcal{V}_{\mathrm{S}} \cup \{\mathbf{v}^\star\}$
\ENDWHILE

\STATE $\mathcal{V}_{\mathrm{keep}} \leftarrow \mathcal{V}_{\mathrm{S}}$

\STATE \textbf{return} $\mathcal{V}_{\mathrm{keep}}$

\end{algorithmic}
\end{algorithm}
Here, $\lambda{=}1$ considers only crossmodal relevance towards user prompts, and $\lambda{=}0$ measures only internal redundancy. 
Compared with consensus, this prominent relevance captures the strongest crossmodal signal ("keywords") and intra-modal redundancy ("discriminative patches").
We justify that two aggregation strategies are in the same optimization direction in App.~\ref{subsec:app_theory_max}, and present 
empirical comparisons in App.~\ref{subsec:exp_ablation}.

\paragraph{Greedy search.} 
Finally, we conduct a greedy search following Eqn.~(\ref{eq:greedy}), to find $\mathbf{v}^\star$ that maximizes marginal gain per step:
\begin{align}
    \mathbf{v}^\star &= \arg\max_{\mathbf{v}_i \in \mathcal{V} \setminus \mathcal{V}_\mathrm{S}} S_i.
\end{align}

\subsection{Efficient Inference}
\label{subsec:methods_eff}
Our method prunes vision tokens $\mathcal{V}\in \mathbb{R}^{N_V \times d}$ to a smaller set $\mathcal{V}_\mathrm{keep}\in \mathbb{R}^{N_\mathrm{keep} \times d}$. Consequently, the input of LLMs is updated to $\mathcal{X}\in \mathbb{R}^{(N_\mathrm{keep}+N_T) \times d}$:
\begin{align}
    \mathcal{X}&= \mathcal{V}_\mathrm{keep}\oplus \mathcal{T} \in \mathbb{R}^{(N_\mathrm{keep}+N_T) \times d}, ~N_\mathrm{keep}<N_V.
\end{align}
Since we don't require attention collection or intervention of LLMs, our MI-based calculation is very efficient with minimal latency. Meanwhile, the subset selection problem becomes \emph{modular} instead of just submodular in our derivation. Following the scoring-and-selection paradigm in general pruning methods \citep{wang2025autoprune,zhang2025cdpruner}, we analyze the computational complexity in these two parts. Please refer to App.~\ref{subsec:app_theory_eff}.

\section{Experiments}
\label{sec:exp}

\subsection{Experimental Setup}
\label{subsec:exp_setup}

\paragraph{Evaluation.}
We conduct experiments on GQA~\citep{hudson2019gqa}, SQA \citep{lu2022learn} (image subset), VQA$_\mathrm{text}$ \citep{singh2019towards} (reference answer provided), Seed \citep{li2024seed} (image subset), MME$_\mathrm{P}$ \citep{fu2025mme} (Yes/No) and POPE \citep{li2023evaluating} (Yes/No). 
More dataset descriptions can be found in App.~\ref{sec:app_datasets}.
The inference and evaluation follow the default settings and metrics. Specifically, we report F1 score (\textbf{P}$_\mathbf{F1}$) and Yes-rate (\textbf{P}$_\mathbf{Yes}$) for POPE as fine-grained measures.
The MLLMs include LLaVA1.5 \citep{liu2024improved}, Qwen2VL \citep{wang2024qwen2}, Qwen2.5VL \citep{Qwen2.5VL} and latest Qwen3VL \citep{Qwen3VL}, all in greedy sampling and SDPA attention. We report main results in Sec.~\ref{subsec:exp_results}, and provide more results in App.~\ref{subsec:app_exp_large}.

\paragraph{Comparison methods.} 
Our comparison involves the baseline with full tokens, salience- and subset-based methods and two naive approaches.
For LLaVA1.5, we compare with \textit{(i) salience-based methods} FastV \citep{chen2024image}, SparseVLM \citep{zhang2024sparsevlm}, VisionZip \citep{yang2025visionzip} and VisPruner \citep{zhang2025beyond}, \textit{(ii) subset-based methods} DART \citep{wen2025stop}, HoloV \citep{zou2025don}, ApeT \citep{ma2026apet}, and \textit{(iii) naive approaches}, where "Random" denotes averaging three independent runs with randomly selected tokens, and "\cosine"~refers to cosine similarity between vision and text tokens (Eqn.~(\ref{eq:sim})). 
As a variation, \mia~combines attention from the vision encoder (round-1) and MI-guided pruning (round-2).
Since existing methods haven't released codes on Qwen3VL, we establish a general "Attention" baseline from VisionZip.

\paragraph{Implementation details.}
All experiments are conducted on a single NVIDIA A100 GPU with batch-size${=}1$. We reproduce other pruning methods from their official repositories. The efficiency results are tested on POPE with 100 runs, see more in Sec.~\ref{subsec:app_exp_efficiency}.

\begin{table*}[h]
\centering
\setlength{\tabcolsep}{5pt} 
\caption{\textbf{Performance comparison of different pruning methods on LLaVA1.5-7B.} We report the latency tested on a single A100 GPU with error bars (from 100 runs).
For POPE, a higher F1 score (\textbf{P$_\mathbf{F1}$}) and a balanced Yes-ratio (\textbf{P$_\mathbf{Yes}$}) around 0.5 indicate better performance.}
\begin{adjustbox}{width=\textwidth}
\begin{tabular}{l ccccc ccc |cc} 
\hline
 {\textbf{Methods}} 
&  {\textbf{GQA}} 
&  {\textbf{SQA}} 
& {\textbf{VQA$_\mathrm{text}$}} 
&  {\textbf{Seed}} 
&  {\textbf{MME$_\mathrm{P}$}} 
& {\textbf{POPE}} 
& {\textbf{P$_\mathbf{F1}$}} 
& {\textbf{P$_\mathbf{Yes}$}}
& {\textbf{Avg$\uparrow$}}
& {\textbf{Latency$\downarrow$}} \\
\hline
\textit{\textbf{LLaVA1.5-7B}} 
& 61.97 & 66.80 & 59.09 &  64.45 & 1509.13 & 85.18 & 0.86 & 0.56 & 67.50 & 106.08$_{\pm 0.34}$ \\
\rowcolor{lightgray}
\multicolumn{11}{c}{\textit{keep 64 (88.9\% $\downarrow$)}} \\
\rowcolor{white}
Random 
& 55.45 & 67.34 & 49.54 &  58.92 & 1274.27 & 79.82 & 0.76 & 0.34 & 	62.21  & - \\
\cosine 
& 54.63 & 68.47 & 50.30 & 59.10 &  1286.66 & 84.24 & 0.83 & \subsota{0.42} & 	63.35 & - \\
FastV \citep{chen2024image} \vvv{ECCV'24}
& 50.80 & 51.10 & 47.80 &  57.82 & 1019.60 & 48.50 & 0.48 & 0.35 & 	51.20 & 104.21$_{\pm 0.30}$ \\
SparseVLM \citep{zhang2024sparsevlm} \vvv{ICML'25}
& 52.70 & 62.20 & 51.80 &  58.61 & 1221.10 & 75.45 & 0.75 & 0.37 & 	60.15 & 93.67$_{\pm 0.31}$ \\
VisionZip \citep{yang2025visionzip} \vvv{CVPR'25}
& 55.10 & 69.00 &  55.50 &  59.53 & 1365.60 & 77.34 & 0.76 & 0.37 & 	63.29 & 89.52$_{\pm 0.29}$ \\
VisPruner \citep{zhang2025beyond} \vvv{ICCV'25}
& 55.40 &  69.10  & \subsota{55.80} &  59.67 &  1369.90  & 82.61 & 0.80 & 0.38 & 	64.52 & 89.48$_{\pm 0.32}$ \\
DART \citep{wen2025stop} \vvv{EMNLP'25}
& 55.90 & 68.86 & 54.40 &  58.73 & 1273.30 & 73.90 & 0.73 & 0.33 & 	62.36 & 88.41$_{\pm 0.37}$ \\
HoloV \citep{zou2025don} \vvv{NeurIPS'25}
& 55.30 & 68.62 & 55.40 & 57.47 & 1367.11 & 80.30 & 0.80 & 0.37 & 	63.42 & 90.12$_{\pm 0.35}$ \\
ApET \citep{ma2026apet} \vvv{CVPR'26}
& 56.85 & 68.90 & 53.00 & 57.45 & 1270.20 & 84.40 & \sota{0.84} & 0.39 & 64.12 & 109.04$_{\pm 0.19}$ \\
\mia
& \sota{57.01} & \subsota{69.20} & \sota{55.90} & \subsota{60.37} & \sota{1428.03} & \sota{85.04} & \sota{0.84} & 0.41 & \sota{65.50} & \subsota{85.12$_{\pm 0.33}$} \\
\mi  
& \subsota{56.88} & \sota{69.81} & 54.91 & \sota{60.78} &  \subsota{1381.35} & \subsota{84.91} & {0.83} & \sota{0.43}  & \subsota{65.46} & \sota{77.06$_{\pm 0.36}$} \\
\rowcolor{lightgray}
\multicolumn{11}{c}{\textit{keep 32 (94.4\% $\downarrow$)}} \\ 
\rowcolor{white}
Random 
& 52.58 & 67.33 & 47.20 & 53.91 &  1176.85 & 73.23 & 0.66 & 0.30 & 	58.85 & - \\
\cosine 
& 51.46 & 67.48 & 47.78 & 54.25 & 1141.68 & 80.07 & 0.78 & 0.42 & 	60.21 & - \\
FastV \citep{chen2024image} \vvv{ECCV'24}
& 41.50 & 42.60 & 42.50 &  53.12 & 884.60 & 33.50 & 0.33 & 0.34 & 42.64 & 101.23$_{\pm 0.29}$ \\
SparseVLM \citep{zhang2024sparsevlm} \vvv{ICML'25}
& 48.30 & 57.30 & 46.10 &  56.80 & 1046.70 & 68.90 & 0.68 & 0.36 & 55.48 & 93.33$_{\pm 0.44}$ \\
VisionZip \citep{yang2025visionzip} \vvv{CVPR'25}
& 51.80 & 68.80 & 53.10 & 55.92 &  1247.40 & 68.92 &  0.68 & 0.37 & 	59.71 & 90.09$_{\pm 0.34}$ \\
VisPruner \citep{zhang2025beyond} \vvv{ICCV'25}
& 52.20 &  \subsota{69.20} & \subsota{53.90} &  56.27 & 1271.00 & 77.83 &  0.73 & 0.32 & 61.88 & 89.98$_{\pm 0.39}$ \\
DART \citep{wen2025stop} \vvv{EMNLP'25}
& 52.77 & 68.76 & 52.20 &  53.45 & 1273.30 & 69.13 & 0.69 & 0.31 & 	59.26 & 87.56$_{\pm 0.47}$ \\
HoloV \citep{zou2025don} \vvv{NeurIPS'25}
& 52.46 & 68.86 & 53.37 & 53.86 & 1268.55 & 81.20 & 0.80 & 0.38 & 61.95 & 80.12$_{\pm 0.27}$ \\
ApET \citep{ma2026apet} \vvv{CVPR'26}
& \subsota{55.12} & 68.92 & 53.54 & 57.12 & 1301.34 & 82.45 & 0.80 & 0.38 & 63.43 & 108.83$_{\pm 0.23}$ \\
\mia
& 54.19 & 69.11 & \sota{54.07}  & \sota{57.75} &  \sota{1344.16} & \subsota{82.81} & \sota{0.81} & \subsota{0.39} & \subsota{63.59} & \subsota{77.36$_{\pm 0.40}$} \\
\mi 
& \sota{55.78} & \sota{69.26} & 52.63 &  \subsota{57.66} & \subsota{1307.72} & \sota{83.18} & \sota{0.81} & \sota{0.41} & 	\sota{63.70} & \sota{76.87$_{\pm 0.32}$} \\
\hline
\end{tabular}
\label{tab:llava_results}
\end{adjustbox}
\end{table*}

\subsection{Main Results}
\label{subsec:exp_results}

\paragraph{LLaVA1.5.}
Following well-established benchmarks \citep{yang2025visionzip,zhang2025beyond}, we report the performance and latency of LLaVA1.5-7B in Tab.~\ref{tab:llava_results}, where \mi~and \mia~deliver SOTA performance on most benchmarks. While DART, another attention-free method, achieves comparable latency to ours, our performance surpasses DART by a significant margin. See more details in "Efficiency analysis".
We additionally report the performance of random pruning, which includes no importance or similarity measure.
Surprisingly, the well-known methods FastV and SparseVLM don't really outperform "Random" as illustrated in \cite{wang2025all,wen2025stop,wen2025token}, underscoring the necessity of a solid theoretical ground. 
Moreover, most comparison methods significantly reduce the proportion of "Yes" response in POPE under 0.40, likely due to mistakenly pruning image tokens corresponding to the queried objects. 
In contrast, our method maintains a more balanced Yes/No distribution by preserving queried information in image tokens. Fig.~\ref{fig:vis_llava} visualizes the pruning effects of representative methods working in the projection space (ours), in LLMs (SparseVLM) and in vision encoders (VisPruner). 
\mi~consistently catches the queried region in three budgets.
Notably, our approach does not require collecting attention maps, making it non-intrusive and more suitable for practical deployment.
\begin{figure}[h]
  \vspace{-0.3cm}
  \centering
  \includegraphics[width=0.85\columnwidth]{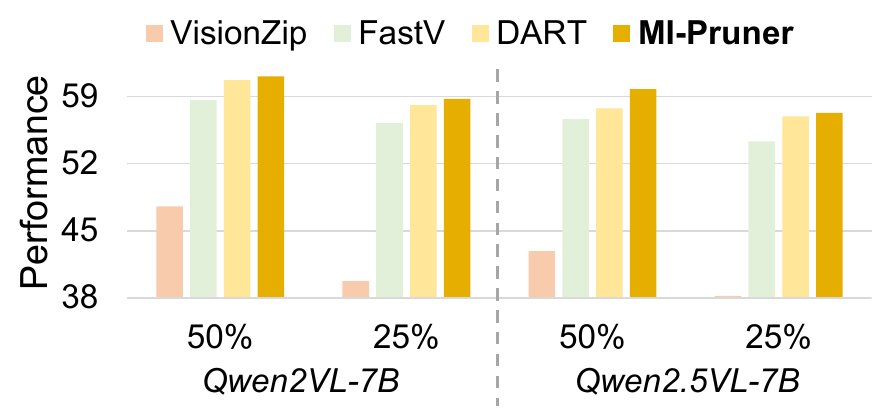}
  \vspace{-0.3cm} 
\caption{\textbf{Performance on Qwen2VL series (GQA).}}
  \label{fig:results_qwen2} 
\end{figure}

\begin{table*}[!h]
\centering
\setlength{\tabcolsep}{5pt} 
\caption{\textbf{Performance comparison of different pruning methods on Qwen3VL-2B and -8B.} "Attention" refers to pruning by the sum of attention scores from other image tokens \citep{yang2025visionzip,zhang2025beyond}. \mi~and \mia~achieve SOTA performance on both scales.}
\begin{adjustbox}{width=\textwidth}
\begin{tabular}{l ccccc|l ccccc}
\hline
{\textbf{Methods}} & {\textbf{GQA}} 
&  {\textbf{MME$_\mathrm{P}$}} 
& {\textbf{POPE}} 
& {\textbf{P$_\mathbf{F1}$}} 
& {\textbf{P$_\mathbf{Yes}$}} 
& {\textbf{Methods}} & {\textbf{GQA}} 
&  {\textbf{MME$_\mathrm{P}$}} 
& {\textbf{POPE}} 
& {\textbf{P$_\mathbf{F1}$}} 
& {\textbf{P$_\mathbf{Yes}$}} \\
\hline

\textit{\textbf{Qwen3VL-2B}} &59.55 & 1494.98   &   89.89 & 0.90 & 0.47   & \textit{\textbf{Qwen3VL-8B}}   &   61.53   & 1730.47  &   88.84 & 0.88 & 0.44  \\
\rowcolor{lightgray}
\multicolumn{12}{c}{\textit{keep 50\%}} \\
\rowcolor{white}
Random    &  49.96  &   1438.19  &  88.76 & 0.88 & 0.46  &   Random    &  53.02 &   1422.31 &  88.25 & 0.87 & 0.43  \\
\cosine   &  50.87   &  1454.34 & 89.26 & 0.89 & 0.48  &  \cosine   & 53.17  &  1461.25  &  88.95 & 0.88 & 0.45  \\
Attention &  50.09 &   1462.74 & 89.54 & 0.89 & 0.47   &  Attention &   52.93 &  1462.84 &  88.85 & 0.88 & 0.44   \\
\mi       &   \sota{51.30}   &   \subsota{1478.26}  &   \sota{89.78} & \sota{0.90} & 0.48 &  \mi       &   \sota{53.48} &  \sota{1493.04} &  \sota{89.22} & \sota{0.89} & 0.45   \\
\mia &    \subsota{51.11} &  \sota{1493.20}  &   \subsota{89.73} &  \subsota{0.89} & 0.48 &  \mia       &   \subsota{53.23} &  \subsota{1482.10} &  \subsota{89.00} & \subsota{0.88} & 0.44    \\
\rowcolor{lightgray}
\multicolumn{12}{c}{\textit{keep 25\%}} \\
\rowcolor{white}
Random    &  47.15   & 1398.52 &  85.59 & 0.85 & 0.42 &   Random    & 49.47  & 1256.18&   85.64 & 0.84 & 0.41  \\
\cosine   &  48.43  & 1372.50  &  88.28 & 0.88 & 0.47 &  \cosine   &  50.61  & 1338.28 &   87.98 & 0.88 & 0.45    \\
Attention & 46.59   & 1158.90  &  86.39 & 0.85 & 0.41 &  Attention &  49.00  & 1293.91 & 87.74 & 0.87 & 0.43      \\
\mi       &  \sota{49.63}  & \subsota{1447.14} & \sota{89.51} & \sota{0.89} & 0.48 &  \mi    &  \sota{51.42}   & \subsota{1369.58}  &  \sota{88.82} & \sota{0.88} & 0.45 \\
\mia  &   \subsota{48.53}  &    \sota{1459.36}  &   \subsota{89.38} & \sota{0.89} & 0.46   &  \mia       &  \subsota{50.72} &  \sota{1415.71} & \sota{88.82} & \sota{0.88} & 0.44   \\
\hline
\end{tabular}
\label{tab:qwen3_results}
\end{adjustbox}
\end{table*}

\begin{figure*}[h]
\centering
\begin{subfigure}{0.49\linewidth}
    \centering
\includegraphics[width=\linewidth]{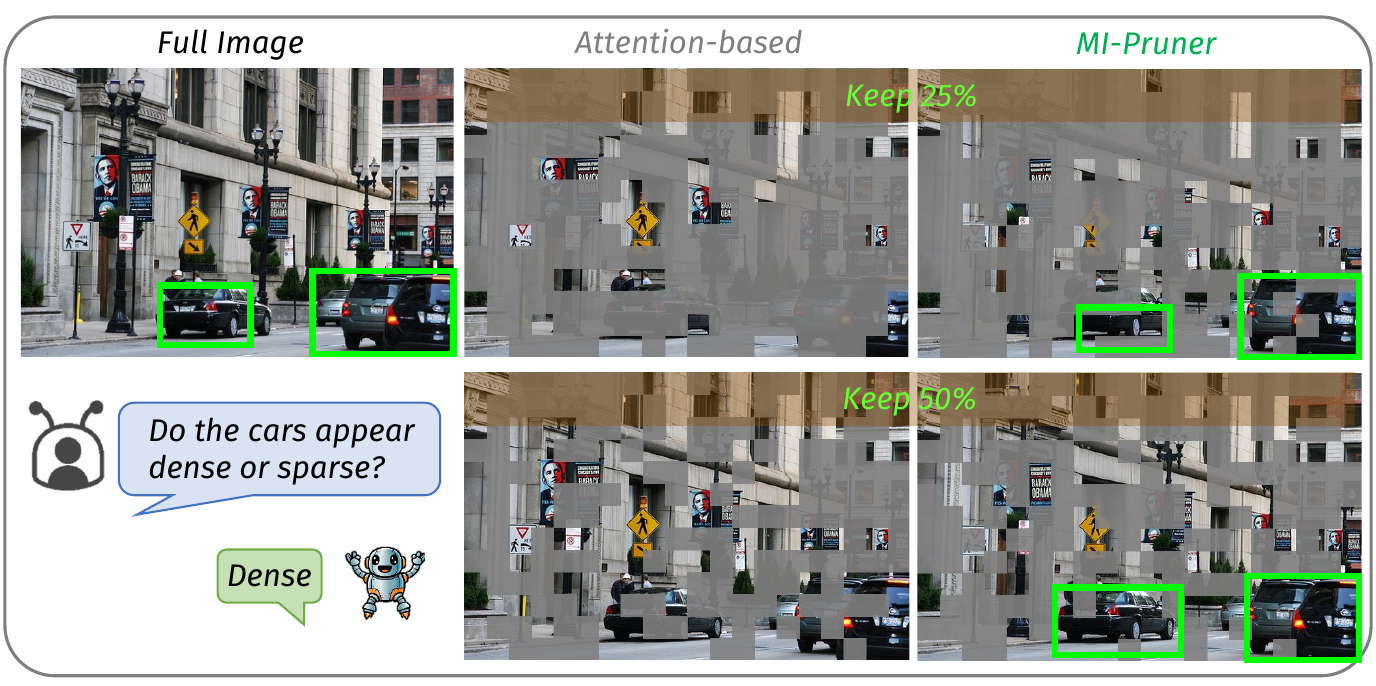}
    \caption{Qwen3VL-2B}
    \label{subfig:vis_qwen3_2b}
\end{subfigure}
\begin{subfigure}{0.49\linewidth}
    \centering
    \includegraphics[width=\linewidth]{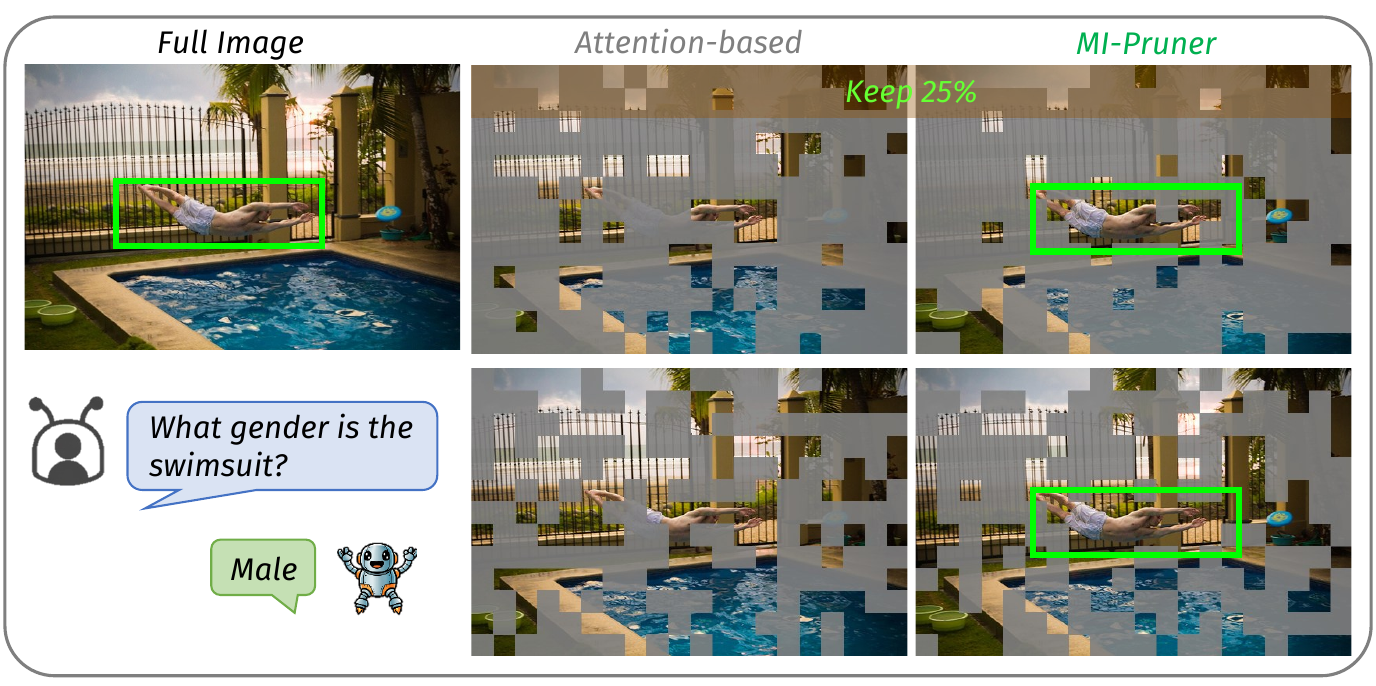}
    \caption{Qwen3VL-8B}
    \label{subfig:vis_qwen3_8b}
\end{subfigure}
\vspace{-0.1cm}
\caption{\textbf{Pruning visualization on Qwen3VL.}
MI-Pruner retains semantically relative patches regarding prompts adaptively, while salience-based methods tend to focus on logos and faces regardless of prompts.}
\label{fig:vis_qwen3}
\end{figure*}

\vspace{-0.5cm}
\paragraph{Qwen2VL and Qwen3VL series.}
Despite the model-agnostic nature, we further evaluate our method on advanced Qwen-series with dynamic resolutions. 
For Qwen2VL series, we show performance on different generations: 
2VL and 2.5VL in Fig.~\ref{fig:results_qwen2}. 
The visual cues in VisionZip depend on specific vision encoders, which limit its generalization capacity. 
In comparison, our method outperforms both subset-based DART and salience-based VisionZip and FastV.
For the latest Qwen3VL series, we 
illustrate two scales 2B and 8B in Tab.~\ref{tab:qwen3_results}, where "Attention" is implemented as the attention sum-up from other image tokens, to avoid the specific design of [\texttt{CLS}] or hyperparameters of existing methods \citep{yang2025visionzip, zhang2025beyond}. Meanwhile, the visualization of pruning effects can be found in Fig.~\ref{fig:vis_qwen3}. By design, high-attention tokens produced by the vision encoder are biased toward generic salient features, such as faces or logos, which are pivotal for classification. However, such prompt-agnostic pruning exhibits reduced robustness when the query lies in non-salient or fine-grained regions that fall outside these pre-defined regions of interest. For instance, the Attention method keeps the facial landmark in Fig.~\ref{subfig:vis_qwen3_2b} while pruning the queried cars. 
The scattered patches among the swimmer and 
the swimming pool in Fig.~\ref{subfig:vis_qwen3_8b} increase the difficulty of multimodal reasoning. 
In comparison, \mi~provides semantically complete patches aligned with the query.

\begin{table}[!h]
\centering
\setlength{\tabcolsep}{5pt} 
        \caption{\textbf{Efficiency comparison on LLaVA1.5-7B.}}
        \label{tab: efficiency_llava}
        \resizebox{0.49\textwidth}{!}{ 
            \begin{tabular}{l |cc |cc} 
            \hline
            \textbf{Methods} & \textbf{Mem (GB)} & \textbf{Latency (ms)} & \textbf{Mem (GB)} & \textbf{Latency (ms)} \\
            \hline
            \rowcolor{lightgray}
              & \multicolumn{2}{c|}{\textit{keep 128}} & \multicolumn{2}{c}{\textit{keep 64}}\\ 
              \rowcolor{white}
            SparseVLM \vvv{ICML'25} &   18.08  &  96.36$_{\pm 0.35}$  &   18.11 & 93.67$_{\pm 0.31}$ \\
            VisPruner \vvv{CVPR'25} & 14.35  &  92.04$_{\pm 0.44}$ & 14.35   & 89.48$_{\pm 0.32}$  \\
            CDPruner \vvv{NeurIPS'25}  & 14.73  & 111.12$_{\pm 0.65}$  &  14.65 & 91.41$_{\pm 0.41}$ \\
            DART \vvv{EMNLP'25} & 14.05 & 92.62$_{\pm 0.39}$  & 13.96  & 88.41$_{\pm 0.37}$ \\   
            \mi$_{\lambda=0.5}$ & 14.01 & 98.92$_{\pm 0.40}$  & 13.93  & 84.95$_{\pm 0.34}$ \\ 
            \mi$_{\lambda=1}$ &  \sota{14.01}   &  \sota{82.08$_{\pm 0.32}$} &  \sota{13.93}   &  \sota{77.06$_{\pm 0.36}$}  \\
            \hline
            \end{tabular}
}
\end{table}

\vspace{-0.2cm}
\paragraph{Efficiency analysis.}
The big-O complexity is analyzed in Sec.~\ref{subsec:methods_eff}. Here, we report the GPU memory and latency on POPE in Tab.~\ref{tab: efficiency_llava}. Since SparseVLM operates in LLMs, its efficiency benefits are suboptimal. Despite the early stage in vision encoder, VisPruner is still held back by its attention collection. As a diversity-based method, CDPruner compromises efficiency due to MAP inference. 
Despite a slight latency increase when intra-model redundancy is considered ($\lambda \neq 1$), \mi~delivers the best inference efficiency under both configurations.

\subsection{Ablation Study}
\label{subsec:exp_ablation}

\paragraph{Aggregation strategies. }
We adopt max aggregation in our scoring function (Eqn.~(\ref{eq:proxy_cross})-(\ref{eq:proxy_self})). Tab.~\ref{tab: ablation_llava} (top) compares the performance of max and average aggregation under different token retention settings (64 and 32 tokens). Empirically, our max aggregation consistently outperforms average aggregation. We hypothesize that max aggregation is more effective at preserving the most discriminative semantic cues associated with the queried object, whereas the average aggregation tends to dilute these salient signals by incorporating less relevant token interactions.

\vspace{-0.2cm}
\paragraph{Hyperparameters. }
The temperature $\tau$ in Eqn.~(\ref{eq:sim}) controls the sharpness of similarity scores. As shown in Tab.~\ref{tab: ablation_llava} (bottom) $\tau{=}0.1$ achieves the best performance on GQA and SQA.
In addition, the factor $\lambda$ in Eqn.~(\ref{eq:score_2}) modulates the scoring function to be crossmodal-only, intra-modal-only or a combination of both. As a task-related parameter, the cross- and intra-modal combination ($\lambda{=}0.5$) suits open-ended datasets (like GQA), and the crossmodal guided pruning is sufficient for closed-form datasets (like SQA) in Tab.~\ref{tab: ablation_llava} (bottom), since the prompt explicitly encodes the ground-truth cue in choices. The highest metrics are annotated for distinguishing. Please find more results of the ablation study in App.~\ref{subsec:exp_ablation_app}.

\begin{table}[h]
\centering
\setlength{\tabcolsep}{5pt} 
\caption{\textbf{Ablation study on LLaVA1.5-7B.} GQA$_{64}$ denotes 64 tokens retained in GQA dataset.}
\label{tab: ablation_llava}
\resizebox{0.48\textwidth}{!}{ 
\begin{tabular}{l cc cc} 
\hline
\textbf{Settings} & \textbf{GQA$_{64}$} & \textbf{GQA$_{32}$}  &\textbf{SQA$_{64}$} & \textbf{SQA$_{32}$}  \\
\hline
\rowcolor{lightgray}
\multicolumn{5}{c}{\textit{aggregation}} \\
\rowcolor{white}
average   & 54.86 & 49.02  &   67.23  &  66.48 \\
max   &  \sota{56.88}  &  \sota{55.78}  &   \sota{69.81}  &  \sota{69.26} \\
\hline
\rowcolor{lightgray}
\multicolumn{5}{c}{\textit{hyperparameters}} \\
\rowcolor{white}
$\tau{=}0.5$, both   &  54.32  &  52.23  &   68.56  & 68.99  \\
$\tau{=}0.1$, intra-only   &  56.59 & 50.83 & 68.76 & 68.21 \\
$\tau{=}0.1$, cross-only   &  56.59 & 50.83 &   \sota{69.81}  &  \sota{69.26} \\
$\tau{=}0.1$, both   &  \sota{56.88}  &  \sota{55.78}  &   68.63 & 68.07 \\
\hline
\end{tabular}
}
\end{table}

\subsection{Video Understanding Results}
\label{subsec:exp_results_video} 
For generalization, we also apply our method to Video-LLaVA-7B \citep{lin2024video} 
and test on TGIF-QA \citep{jang2017tgif}, MSVD-QA \citep{xu2017video} and MSRVTT-QA \citep{xu2017video} datasets.
Following previous work \citep{yang2025visionzip}, we report video-QA pruning results on Video-LLaVA \citep{lin2024video}, which processes 8 frames with $256\times8{=}2048$ visual tokens. Tab.~\ref{tab:video}, compares ours with FastV and VisPruner under 227 (11.08\% retained) and 114 (5.57\% retained) video tokens budgets, scored by GPT-3.5-Turbo \citep{openai2023gpt35turbo}.
\mi~achieves average 93.19\% accuracy across three datasets with merely 114 visual tokens, which indicates our robustness on video pruning.

\begin{table}[!h]
\centering
\setlength{\tabcolsep}{5pt} 
        \caption{\textbf{Performance on Video-LLaVA-7B.} \mi~maintains good performance on video understanding tasks.}
        \label{tab:video}
        \resizebox{0.49\textwidth}{!}{ 
            \begin{tabular}{l cccccc} 
            \hline
            \multirow{2}{*}{\textbf{Methods}} 
            & \multicolumn{2}{c}{\textbf{TGIF}} & \multicolumn{2}{c}{\textbf{MSVD}} & \multicolumn{2}{c}{\textbf{MSRVTT}}\\
            & \textit{Acc } & \textit{Score } & \textit{Acc} & \textit{Score} & \textit{Acc } & \textit{Score} \\
            \hline
            \textit{\textbf{Video-LLaVA-7B}} 
            & 18.90 & 2.54 & 72.00 & 3.95 & 57.50 & 3.50 \\
            \rowcolor{lightgray}
            \multicolumn{7}{c}{\textit{keep 227}} \\ 
             \rowcolor{white}
            FastV \vvv{ECCV'24} & 14.30 & 2.42 & 68.90 & 3.90 & 53.00 &  3.40 \\
            VisPruner \vvv{ICCV'25} &  \sota{15.90} & 2.41 & 69.30 & 3.92 & 55.60 & 3.45 \\
            \mi &  15.70 & \sota{2.44} & \sota{70.60} & \sota{3.94} & \sota{56.40} & \sota{3.46} \\
            \rowcolor{lightgray}
            \multicolumn{7}{c}{\textit{keep 114}} \\ 
            \rowcolor{white}
            FastV \vvv{ECCV'24}& 10.60 & 2.29 & 64.10 & 3.78 & 52.40 & 3.39 \\
            VisPruner \vvv{ICCV'25} &  \sota{14.10} & 2.35 & 65.40 & 3.79 & 54.10 & 3.41 \\
            \mi &  13.10 & \sota{2.40} & \sota{69.90} & \sota{3.92} & \sota{55.30} & \sota{3.49} \\
            \hline
            \end{tabular}
        }
\end{table}

\section{Discussions}
\label{sec:exp_discuss}

\paragraph{Relationship with attention mechanism. }
The inner product and softmax operations in \mi~also appear in the attention mechanism \citep{vaswani2017attention}.
However, our algorithm differs from the attention mechanism in three aspects.
\textit{(i) Implementation:} Our scoring function is conducted just once in the projection space, after which the attention burden in LLM blocks is largely reduced. 
\textit{(ii) Training-free nature:} We require no training step to learn the QKV matrices as the attention module; therefore, \mi~is versatile and ready to be applied to any pretrained MLLMs. 
\textit{(iii) Theoretical ground:} Our importance measure is derived from Mutual Information \citep{jaynes1957information}, which is a traditional metric in information theory (details in App.~\ref{subsec:pre_mi}) to quantify the statistical dependence among tokens. Framing the token selection issue in a submodular framework, \mi~achieves SOTA performance from 
information-theoretic motivation
, not from a scaling law \citep{vaswani2017attention}.

\paragraph{Applications and benefits. } 
Once conducted in the projection space, we decrease the visual attention cost in each LLM block from $N_V^2$ to $N_\mathrm{keep}^2$, where $N_\mathrm{keep}\ll N_V$. 
For instance, LLaVA1.5-7B \citep{liu2024improved} employs 32 LLM blocks and 576 image tokens. The visual attention computation scales with $32\cdot576^2$ by default, but can be decreased to $32\cdot32^2$ (9 million$^\downarrow$) in our 32 setting.
Experiments validate our performance (Tab.~\ref{tab:llava_results} and~\ref{tab:qwen3_results}) and efficiency (Tab.~\ref{tab: efficiency_llava}). Moreover, \mi~ is easy to implement and supports various attention cores, like FlashAttention (see Tab.~\ref{tab: app_efficiency} in appendix).
Rooted in submodularity, our MI-based token selection is theoretically grounded and offers enhanced interpretability. Unlike previous visual-only token reduction methods \citep{zhang2025beyond,yang2025visionzip}, MI-Pruner explicitly accounts for both crossmodal and intra-modal dependencies, thereby demonstrating superior robustness. 
The visualization presented in Fig.~\ref{fig:vis_llava} and Fig.~\ref{fig:vis_qwen3} validates the interpretability of our approach over existing baselines.

\section{Conclusion}
\label{sec:conclusion}

We propose \mi, a plug-in and attention-free visual pruning method. Formulating token pruning as a subset selection problem, we propose a relevance score based on Mutual Information.
Without relying on attention maps, \mi~works surgically in the projection space, providing a principled and efficient framework for visual token reduction. Experiments and visualization verified our robustness and efficiency. The non-intrusive manner enables our method to be applied to any off-the-shelf MLLMs with the most favorable inference speedup.

\paragraph{Limitations.} 
To make it tractable, we assume equal visual probability and conditional independence in PMI estimation.
While a more informative visual prior may improve the performance, obtaining an appropriate prior is non-trivial and remains an open challenge \citep{chen2026priortr,wang2025autoprune}. Please refer to App.~\ref{subsec:app_exp_more} for more discussions. 

\clearpage

%% file: sec/3_appendix.tex
\clearpage

\appendix

\twocolumn[
   \begin{center}
      \LARGE \textbf{Appendix}
   \end{center}
]

\section{Theory}
\label{sec:app_theory}

\subsection{Definition}
\label{subsec:app_theory_def}

\begin{definition}[Shannon Entropy \citep{kraskov2004estimating,gray2011entropy,gallager1968information}]
The Shannon entropy $H(X)$ of a random variable $X$ measures the average uncertainty or information content associated with $X$:
\begin{equation}
    H(X) = -\sum_{x \in \mathcal{X}} p(x) \log p(x). 
\end{equation}
\end{definition}

\begin{definition}[Conditional Entropy \citep{kraskov2004estimating,gray2011entropy,gallager1968information}]
Given two random variables $X$ and $Y$, the conditional entropy $H(X|Y)$ quantifies the amount of information needed to describe $X$ given that known $Y$:
\begin{align}
    H(X|Y) 
       &= -\sum_{u \in \mathcal{X}, y \in \mathcal{Y}} p(x, y) \log p(x|y).
\end{align}
\end{definition}

\begin{definition}[Conditional Mutual Information \citep{gallager1968information}]
\label{def:cmi}
The Conditional Mutual Information $\mathrm{MI}(X; Y \mid Z)$ measures the reduction in uncertainty of a random variable $X$ given $Z$ when additionally knowing $Y$:
\begin{align}
    \mathrm{MI}(X; &Y \mid Z) 
    = H(X \mid Z) - H(X \mid Y, Z) \\
    &= \sum_{x \in \mathcal{X},\, y \in \mathcal{Y},\, z \in \mathcal{Z}} 
p(x, y, z) \log \frac{p(x \mid y, z)}{p(x \mid z)}.
\end{align}
\end{definition}

\begin{definition}[Submodular Function \citep{fujishige2005submodular}]
Let $X$ be a finite set. A set function $f: 2^X \to \mathbb{R}$ is called submodular if for every $A, B \subseteq X$, we have:
\begin{equation}
    f(A \cup B) + f(A \cap B) \leq f(A) + f(B).
\end{equation}
\end{definition}
An equivalent and often more intuitive definition is based on the \textit{marginal gain}, as shown in the main paper.

\subsection{Mutual Information}
\label{subsec:pre_mi}

Information theory provides a rigorous mathematical framework for quantifying information transmission and representation, in which Mutual Information serves as a natural measure of how much information is shared between variables. Applied to MLLMs, MI is able to characterize the statistical dependence among tokens, offering a principled objective for identifying important and irreplaceable tokens.

\begin{definition}[Mutual Information \citep{kraskov2004estimating,gray2011entropy}]
\label{def:mi}
The Mutual Information $\mathrm{MI}(X; Y)$ measures the gap between entropy $H(X)$ and conditional entropy $H(X|Y)$ (see App.~\ref{subsec:app_theory_def} for details), which quantifies the uncertainty reduction of a random variable (r.v.) $X$ given the knowledge of $Y$:
\begin{align}
    \operatorname{MI}(X; Y) &= H(X) - H(X \mid Y) 
    \label{eq:mi_1}
    \\
    &= \sum_{x \in \mathcal{X},\, y \in \mathcal{Y}} p(x, y)
    \underbrace{\log \frac{p(x \mid y)}{p(x)}}_{\operatorname{PMI}(x; y)}.
    \label{eq:mi_2}
\end{align}
\end{definition}

$\operatorname{MI}(X; Y)$ can be interpreted as the weighted average of Pointwise Mutual Information $\operatorname{PMI}(x, y)$ under the joint distribution $p(x, y)$.
Replacing $x,y$ with embeddings $\mathbf{v}_i, \mathbf{t}_j$ establishes a link to multimodal token relevance. 
In general, maximizing MI is computationally intractable. However, under the assumption\footnote{This assumption is necessary to make MI tractable, and has been adopted in prior studies employing greedy search (see Related Work).} of conditional independence (the components of $X$ are independent given $Y$), it becomes a cardinality-constrained monotone submodular maximization problem \citep{krause2008near}.
This is a well-studied problem with a known approximation guarantee of $1{-}1/e$.

\subsection{Information Theory} 
\label{subsec:app_theory_sub}

In the following, we introduce the marginal gain and diminishing returns of submodular functions.
\begin{definition}[Marginal Gain \citep{fujishige2005submodular}]
\label{def:delta_f}
Let $\mathcal{X}$ be a ground set, and let 
$f: 2^{\mathcal{X}} \to \mathbb{R}$ be a set function.
The marginal gain of adding an element $i \in \mathcal{X} \setminus \mathcal{A}$
to a selected set $\mathcal{A} \subseteq \mathcal{X}$ is defined as:
\begin{equation}
\Delta f(i | \mathcal{A}) = f(\mathcal{A} \cup \{i\}) - f(\mathcal{A}).
\label{eq:delta_f}
\end{equation}
\end{definition}

\begin{definition}[Submodularity and Diminishing Returns \citep{fujishige2005submodular}]
\label{def:sub}
A function $f$ is submodular if and only if for all $\mathcal{A} \subseteq \mathcal{B} \subseteq \mathcal{X}$ and $i \in \mathcal{X} \setminus \mathcal{B}$:
\begin{align}
\Delta f(i | \mathcal{A}) &\geq \Delta f(i | \mathcal{B}).
\label{eq:diminish_1}
\end{align}
\end{definition}

The \textbf{\textit{monotonicity}} motivates our MI-guided visual selection.
Let $\mathcal{V}_\mathrm{S} \subseteq \mathcal{V}$ be a selected set of a ground set of variables, and let $\mathcal{T}$ be a targeted set of variables, we define  Mutual Information as a set function:
\begin{align}
f(\mathcal{V}_\mathrm{S}) &= \mathrm{MI}(\mathcal{V}_\mathrm{S}; \mathcal{T}).
\label{eq:mi_cross}
\end{align}
Following Eqn.~(\ref{eq:delta_f}), the marginal gain of adding a new token $\mathbf{v}_i$ into $\mathcal{V}_\mathrm{S}$ is expressed as:
\begin{align}
\Delta f(\mathbf{v}_i \mid \mathcal{V}_{\mathrm{S}})
&= \mathrm{MI}(\mathcal{V}_{\mathrm{S}} \cup \{\mathbf{v}_i\}; \mathcal{T}) 
- \mathrm{MI}(\mathcal{V}_{\mathrm{S}}; \mathcal{T})
\label{eq:delta_mi_minus} 
\\
&= \mathrm{MI}(\mathbf{v}_i; \mathcal{T} \mid \mathcal{V}_{\mathrm{S}}).
\label{eq:delta_mi_cond} 
\end{align}
The definition of conditional MI is given in App.~\ref{subsec:app_theory_def} (Def.~\ref{def:cmi}), and the proof from Eqn.~(\ref{eq:delta_mi_minus}) to (\ref{eq:delta_mi_cond}) can be found in App.~\ref{subsec:app_theory_cond_indep}.

The cardinality-constrained monotone submodular maximization problem is typically addressed by a greedy search strategy, which selects elements that maximize the marginal gain per step \citep{Nemhauser1978Submodular}:
\begin{align}
    \mathbf{v}^\star &= \arg\max_{\mathbf{v}_i \in \mathcal{V} \setminus \mathcal{V}_\mathrm{S}} \Delta f(\mathbf{v}_i | \mathcal{V}_\mathrm{S}).
    \label{eq:greedy}
\end{align}
When selecting $N_\mathrm{keep}$ from $N_V$ elements, the greedy search holds a complexity of $\mathcal{O}(N_V N_\mathrm{keep})$. In practice, greedy search is a common optimization strategy in token pruning \citep{alvar2025divprune,zhang2025cdpruner}. Building upon it, we develop a selection mechanism with even lower complexity.

\subsection{Conditional Independence} 
\label{subsec:app_theory_cond_indep}
\begin{takeaway}
\textbf{Assumption:} The candidate $\mathbf{v}_i$ is independent of the selected set $\mathcal{V}_\mathrm{S}$ conditioned on $\mathcal{T}$.

\textit{This assumption helps to maintain submodularity and derive tractable marginal gains.}
\end{takeaway}

Mutual Information is widely recognized for exhibiting submodularity (the property of diminishing returns) \citep{iyer2021submodular,krause2008near}, providing a principled foundation for efficient subset selection. 

To rigorously adapt MI-based submodular optimization to the token selection task, we bring the Naive Bayes assumption as the conditional independence between $\mathbf{v}_i$ and $\mathcal{V}_\mathrm{S}$ based on $\mathcal{T}$. From the view of probability and Mutual Information, it holds:
\begin{align}
    p(\mathbf{v}_i, \mathcal{V}_\mathrm{S} \mid \mathcal{T})&=p(\mathbf{v}_i \mid \mathcal{T}) p(\mathcal{V}_\mathrm{S}  \mid \mathcal{T}),
    \label{eq:cond_p}\\
    \mathrm{MI}(\mathbf{v}_i; \mathcal{V}_\mathrm{S} \mid \mathcal{T})&=0.
    \label{eq:cond_mi}
\end{align}
This assumption has two immediate consequences:
(i) \textit{submodular} guarantees for the property of diminishing returns; (ii) a tractable derivation of the \textit{marginal gain}.

\textbf{Submodularity.} The conditional independence assumption assumes that a single patch is sufficiently representative of a semantic unit.
A counterexample is that, if a semantic concept strictly requires the joint presence of multiple patches (e.g., both a left and a right part), then adding the remaining patch at a later stage could violate the diminishing-returns property. Nevertheless, the assumption of semantic sufficiency is well justified in practice. For example, a single patch depicting a “wheel” or a “grille” is often sufficient to convey the semantic concept of “\texttt{\_car}”.
Based on this assumption, we can robustly treat the visual token selection process as a submodular maximization problem, ensuring both theoretical consistency and computational efficiency.

\textbf{Marginal gains.} Since accurately estimating the joint distribution involving $\mathcal{V}_\mathrm{S}$
is high-dimensionally infeasible, we approximate the conditional Mutual Information by the marginal Mutual Information.
According to Def.~\ref{def:mi},
Mutual Information can be expressed in terms of entropy:
\begin{align}
    \mathrm{MI}(X; Y) = H(Y) - H(Y \mid X).
\end{align}
Here, we give the proof from Eqn.~(\ref{eq:delta_mi_minus}) to Eqn.~(\ref{eq:delta_mi_cond}):
\begin{align}
    &\mathrm{MI}(\mathcal{V}_{\mathrm{S}} \cup \{\mathbf{v}_i\}; \mathcal{T}) - \mathrm{MI}(\mathcal{V}_{\mathrm{S}}; \mathcal{T}) \\
    &= \left[ \cancel{H(\mathcal{T})} - H(\mathcal{T} \mid \mathcal{V}_{\mathrm{S}} \cup \{\mathbf{v}_i\}) \right] - \left[ \cancel{H(\mathcal{T})} - H(\mathcal{T} \mid \mathcal{V}_{\mathrm{S}}) \right] \\
    &= H(\mathcal{T} \mid \mathcal{V}_{\mathrm{S}}) - H(\mathcal{T} \mid \mathcal{V}_{\mathrm{S}} \cup \{\mathbf{v}_i\}) \\
    &= \mathrm{MI}(\mathbf{v}_i; \mathcal{T} \mid \mathcal{V}_{\mathrm{S}}).
\end{align}
Further, this conditional MI can be written as:
\begin{align}
&\operatorname{MI}(\mathbf{v}_i; \mathcal{T} \mid \mathcal{V}_{\mathrm{S}})
= \underbrace{\operatorname{MI}(\mathbf{v}_i; \mathcal{T}, \mathcal{V}_{\mathrm{S}})}_{\operatorname{MI}^{\mathrm{cross}}}
 - \underbrace{\operatorname{MI}(\mathbf{v}_i; \mathcal{V}_{\mathrm{S}})}_{\operatorname{MI}^{\mathrm{self}}}\\
 &=\mathrm{MI}(\mathbf{v}_i; \mathcal{V}_{\mathrm{S}} \mid \mathcal{T})+\mathrm{MI}(\mathbf{v}_i;  \mathcal{T})-\operatorname{MI}(\mathbf{v}_i; \mathcal{V}_{\mathrm{S}}).
 \label{eq:mi_late}
\end{align}

However, the crossmodal term is intractable, which causes trouble in deriving the marginal gain $\Delta f$. Again, we leverage the conditional independence assumption, and substitute Eqn.~(\ref{eq:cond_mi}) into Eqn.~(\ref{eq:mi_late}). Since $\cancel{\mathrm{MI}(\mathbf{v}_i; \mathcal{V}_{\mathrm{S}} \mid \mathcal{T})}=0$, we obtain an estimator:
\begin{align}
    \widehat{\mathrm{MI}}(\mathbf{v}_i; \mathcal{T} \mid \mathcal{V}_{\mathrm{S}}) &= \mathrm{MI}(\mathbf{v}_i; \mathcal{T})-\mathrm{MI}(\mathbf{v}_i;  \mathcal{V}_{\mathrm{S}}).
\end{align}
The assumption $\mathrm{MI}(\mathbf{v}_i; \mathcal{V}_{\mathrm{S}} \mid \mathcal{T})=0$ implies that the crossmodal relevance is much stronger than the internal one, \textit{i.e.}, $ \mathrm{MI}^\mathrm{cross} \rightarrow\mathrm{MI}(\mathbf{v}_i, \mathcal{T})$. It is empirically justified under high pruning rates (less than 50\% retained), where sparse selection leads to trivial intra-modal overlap.

For practical implementation, we allow balancing the crossmodal term and the intra-model term with a factor $\lambda$. As shown in Eqn.~(\ref{eq:score_1}), we derive two scoring functions respectively and combine them with $\lambda$.

\subsection{Marginal Distribution} 
\label{subsec:app_theory_prior}
\begin{takeaway}
\textbf{Assumption:} Each image patch occurs with equal probability $p(\tilde{\mathbf{v}}_i)=\frac{1}{N_V}$.

\textit{This assumption helps to derive the marginal text probability $p(\tilde{\mathbf{t}}_j)$.}
\end{takeaway}

In this section, we recap the Law of Total Probability \citep{jaynes1957information} and then derive $p(\tilde{\mathbf{t}}_j)$ in Eqn.~(\ref{eq:margin_t}).
\begin{proposition}[Law of Total Probability \citep{jaynes1957information}]
Let $A$ be an event and let $\{B_i\}_{i=1}^N$ be a set of mutually exclusive and exhaustive events.  
The Law of Total Probability indicates:
\begin{align}
p(A) = \sum_{i=1}^{N} p(A \mid B_i) \, p(B_i).
\label{eq:total_prob}
\end{align}
\end{proposition}
Since the mapping between a patch and a token is deterministic, $\{\tilde{\mathbf{v}}_i\}_{i=0}^{N_V}$ are exclusive and exhaustive. We  derive the marginal probability by substituting $A,B_i$ with $\tilde{\mathbf{t}}_j, \tilde{\mathbf{v}}_i$:
\begin{align}
p(\tilde{\mathbf{t}}_j) &= \sum_{i=1}^{N_V} p(\tilde{\mathbf{t}}_j \mid \tilde{\mathbf{v}}_i) \, p(\tilde{\mathbf{v}}_i).
\label{eq:margin_t_app}
\end{align}
Assuming that $N_V$ patches are equally probable, \textit{i.e.}, the prior probability $p(\tilde{\mathbf{v}}_i)=\frac{1}{N_V}$, Eqn.~(\ref{eq:margin_t}) is formulated as:
\begin{align}
p(\tilde{\mathbf{t}}_j) &= \sum_{i=1}^{N_V} p(\tilde{\mathbf{t}}_j \mid \tilde{\mathbf{v}}_i) \underbrace{\frac{1}{N_V}}_{p(\tilde{\mathbf{v}}_i)}.
\label{eq:margin_t_full}
\end{align}
According to the Principle of Maximum Entropy \citep{gray2011entropy}, our uniformity is the most unbiased assumption, ensuring that the selection process is driven solely by the observed Mutual Information rather than predefined spatial heuristics.
It is non-trivial to obtain the appropriate prior for vision tokens. We admit potential solutions of constructing an estimator from the validation set, or linking with the patch position based on photography composition techniques (\textit{e.g.}, Rule of Thirds and Golden Ratio). 
For clarity, we present a simple ablation study on "central object" prior in App.~\ref{subsec:app_exp_more}.

\subsection{Maximal Aggregation} 
\label{subsec:app_theory_max}

Our max aggregation over $\mathcal{T}$ and $\mathcal{V}_{\mathrm{select}}$ is motivated by the LogSumExp (LSE) operator, defined as
\begin{align}
\mathrm{LSE}(z_1,\dots,z_N) := \log \left( \sum_{j=1}^{N} \exp(z_j) \right).
\end{align}
As a smooth upper approximation to the maximum, the LSE operator satisfies the following bounds:
\begin{align}
    \max_j z_j \;\leq\; \mathrm{LSE}(z_1,\dots,z_N)
    \;\leq\; \max_j z_j + \log N.
    \label{eq:lse_0}
\end{align}

Here, we justify the maximal aggregation as a robust proxy for the overall crossmodal relevance. For simplification, we omit $\tilde{\cdot}$ of $\mathbf{v}_i,\mathbf{t}_j$ Theoretically, let $\mathbf{z}_{j} = \text{PMI}(\mathbf{v}_i; \mathbf{t}_j)$ be the log-likelihood ratio for each token pair:
\begin{align}
\mathbf{z}_j&=\log\frac{p(\mathbf{t}_j\mid \mathbf{v}_i)}{p(\mathbf{t}_j)}.
    \label{eq:lse_1}
\end{align}

The relationship between the peak signal and the total likelihood ratio can be characterized by the LSE operator:
\begin{align}
    \mathrm{LSE}(\mathbf{z}_j)&=\log \left(\sum_{j=1}^{N_T}\exp(\log\frac{p(\mathbf{t}_j\mid \mathbf{v}_i)}{p(\mathbf{t}_j)})\right)
    \\
    &=\log \left(\sum_{j=1}^{N_T} \frac{p(\mathbf{t}_j\mid \mathbf{v}_i)}{p(\mathbf{t}_j)}\right).
    \label{eq:lse_2}
\end{align}
Substitute Eqn.~(\ref{eq:lse_2}) into Eqn.~(\ref{eq:lse_0}), we obtain the lower bound of LSE, and exponentiate both sides:
\begin{align}
    \max_j(\log\frac{p(\mathbf{t}_j\mid \mathbf{v}_i)}{p(\mathbf{t}_j)})  &\leq \log \left(\sum_{j=1}^{N_T} \frac{p(\mathbf{t}_j\mid \mathbf{v}_i)}{p(\mathbf{t}_j)}\right),
    \label{eq:lse_3}\\
    \max_j(\frac{p(\mathbf{t}_j\mid \mathbf{v}_i)}{p(\mathbf{t}_j)})  &\leq  \sum_{j=1}^{N_T} \frac{p(\mathbf{t}_j\mid \mathbf{v}_i)}{p(\mathbf{t}_j)}.
    \label{eq:lse_4}
\end{align}
It's also expressed as:
\begin{align}
        \max_j\left(\mathrm{PMI}(\mathbf{v}_i;\mathbf{t}_j)\right)  &\leq  \sum_{j=1}^{N_T} \mathrm{PMI}(\mathbf{v}_i;\mathbf{t}_j).
    \label{eq:lse_5}
\end{align}
Maximal aggregation serves as a tight lower bound for the sum of PMI, which has the same optimization direction as the global aggregation. 
In practice, crossmodal correspondence exhibits inherent sparsity: within a sentence, typically only a few core tokens are semantically grounded in a specific visual region. Therefore, averaging these sparse, high-intensity signals with numerous irrelevant tokens leads to a noisy dilution of the relevance signal. In contrast, the max aggregation, analogous to an $L_\infty$-type norm of the association scores, captures the most salient semantic tokens and preserves the alignment measure from noise.

\subsection{Efficiency Analysis} 
\label{subsec:app_theory_eff}

\paragraph{Scoring complexity.} 
The scoring function in Eqn.~(\ref{eq:score_2}) consists of two terms.
The crossmodal term holds a complexity of $\mathcal{O}(N_V N_T)$, \textit{i.e.}, all text tokens attend the scoring process.
Yet, the aggregation of intra-modal term is based on the increasing tokens in the selected set $\mathcal{V}_\mathrm{S}$, instead of all visual tokens in $\mathcal{V}$. Therefore, the intra-modal complexity holds  $\mathcal{O}(N_V  N_\mathrm{keep})$, instead of $\mathcal{O}(N_V N_V)$. 
The overall complexity considering two terms is $\mathcal{O}(N_V \cdot \max(N_T,N_\mathrm{keep}))$, and becomes $\mathcal{O}(N_V  N_T)$ when crossmodal priority (${\lambda{=}1}$).

 \paragraph{Selection complexity.}
The selection step aims to choose the top $N_{\mathrm{keep}}$ elements from $N_V$ candidates. 
A naive greedy implementation selects one element per step by scanning all $N_V$ candidates to find the maximum gain, 
resulting in a complexity of $\mathcal{O}(N_V  N_{\mathrm{keep}})$.
In contrast, we maintain a heap over all $N_V$ candidates for efficiency: 
building the heap by a cost of $\mathcal{O}(N_V)$, then performing $N_{\mathrm{keep}}$ retrievals,
each of which costs $\mathcal{O}(\log N_V)$.
Our heap selection complexity yields an overall complexity of $\mathcal{O}(N_V+N_\mathrm{keep}\log N_V)$.
We expect $N_\mathrm{keep}\log N_V<N_V$ in settings of interest, implying a selection complexity of $\mathcal{O}(N_V)$.

\section{More Experiments}
\label{sec:app_exp}

\subsection{Ablation Study}
\label{subsec:exp_ablation_app}

\paragraph{PMI estimation.} We further provide a component-level ablation of the proposed scoring function, \textit{e.g.}, plain cosine \textit{vs} PMI and MinMax \textit{vs} softmax normalization. 
Based on Boltzmann distributions, we compute similarity matrices and apply softmax in Eqn.~(\ref{eq:softmax}). In the appendix, we conduct an ablation study to investigate the impact of linear normalization. Specifically, we implement a two-stage MinMax normalization as follows:
\begin{align}
&\hat{\bm{\rho}}_{ij}^\mathrm{s}=\operatorname{MinMax}_j(\bm{\rho}_{ij}^\mathrm{s})=\frac{\bm{\rho}_{ij}^\mathrm{s}-\operatorname{Min}_j(\bm{\rho}_{ij}^\mathrm{s})}{\operatorname{Max}_j(\bm{\rho}_{ij}^\mathrm{s})-\operatorname{Min}_j(\bm{\rho}_{ij}^\mathrm{s})},
\label{eq:minmax_1}
\\
&p(\tilde{\mathbf{x}}_j \mid \tilde{\mathbf{v}}_i) =\operatorname{Normalize}_j(\hat{\bm{\rho}}_{ij}^\mathrm{s})= \frac{\hat{\bm{\rho}}_{ij}^\mathrm{s}}{\sum_{k=1}^{N} \hat{\bm{\rho}}_{ij}^\mathrm{s}}.
\label{eq:minmax_2}
\end{align}
Here, $\hat{\bm{\rho}}_{ij}^\mathrm{s}$ first maps the raw similarity scores to the $[0, 1]$ range, then Eqn.~(\ref{eq:minmax_2}) ensures a valid probability distribution across the dimension $j$.
As shown in Tab.~\ref{tab: results_plain}, our softmax normalization outperforms MinMax and maintains its overall robust performance. We attribute this performance gap to the non-linear saliency of softmax and its robustness to outliers.
Meanwhile, the "cosine+TopK" setting directly selects visual tokens with the highest crossmodal cosine similarity, which is called “Similarity” in Tab.~\ref{tab:llava_results} and Tab.~\ref{tab:qwen3_results}. A new setting "cosine+greedy" replaces the PMI estimator with plain cosine in Eqn.~(\ref{eq:score_2}) for greedy search.
As observed, ours achieves the best performance on both models by adopting the PMI estimator and softmax normalization.

\begin{table}[h]
\centering
\setlength{\tabcolsep}{5pt} 
\caption{Ablation study on PMI estimation.}
\begin{tabular}{l cccc}
\hline
{\textbf{Methods}} & {\textbf{GQA}} 
&  {\textbf{SQA}} 
& {\textbf{VQA$_\mathrm{text}$}} 
& {\textbf{POPE}} \\
\hline
\rowcolor{lightgray}
\multicolumn{5}{c}{\textit{LLaVA1.5-7B/keep 64}} \\
\rowcolor{white}
MinMax  &  54.79 &  67.90 &  53.23 & 83.31 
\\
cosine+TopK  &  54.61 &  68.54 &  50.35 &  84.24 
\\
cosine+greedy  &  56.02 &  67.89 &  54.23 &  83.92 
\\
\textbf{MI-Pruner} &  \sota{56.88} &  \sota{69.81} &  \sota{54.91} &  \sota{84.91}  
\\
\hline
\rowcolor{lightgray}
\multicolumn{5}{c}{\textit{Qwen3VL-8B/keep 25\%}} \\
\rowcolor{white}
MinMax  & 51.06  &  83.21  &  72.95 &  88.05
\\
cosine+TopK  & 50.64 & 83.92 & 73.19 & 88.01
\\
cosine+greedy  &50.73 & 83.51 & 72.77 & 87.94
\\
\textbf{MI-Pruner} &  \sota{51.42} & \sota{84.02} & \sota{73.51} & \sota{88.82}
\\

\hline
\end{tabular}
\label{tab: results_plain}
\end{table}

\subsection{Large-scale Models and Diverse Settings}
\label{subsec:app_exp_large}
To further demonstrate the robustness of our method, we report results on larger-scale models, including LLaVA1.5-13B and Qwen3VL-30B (Mixture-of-Experts). In Tab.~\ref{tab: app_large_scale}, our method consistently outperforms random sampling and similarity-based pruning. 
Yet, we acknowledge that for the latest models \citep{Qwen3VL}, model-specific tailoring is indispensable. Qwen3VL employs deep-stack integration and adaptive padding mechanisms, which demand dedicated designs to maintain structural integrity and performance during pruning. By providing a principled foundation, our method facilitates future research and remains highly adaptable to next-generation multimodal architectures.
Finally, Tab.~\ref{tab: results_mi} shows a comparison with other MI-based methods.

\begin{table}[h]  
        \centering
        \setlength{\tabcolsep}{5pt} 
        \caption{Results on large-scale models, LLaVA1.5-13B and Qwen3VL-30B.}
        \label{tab: app_large_scale}
        \resizebox{0.49\textwidth}{!}{ 
            \begin{tabular}{l cccc} 
            \hline
            \textbf{Methods} & \textbf{GQA} & \textbf{SQA} & \textbf{VQA$_\mathrm{text}$} & \textbf{MME$_\mathrm{P}$}\\
            \hline
            \textit{\textbf{LLaVA1.5-13B}} & 61.97 & 72.73  & 61.24 & 1524.19 \\
            \rowcolor{lightgray}
            \multicolumn{5}{c}{\textit{keep 64}} \\ 
            \rowcolor{white}
            Random &  53.45 & 70.85  & 50.74   & 1257.35 \\
            cosine &   53.63 & 71.64 & 51.58 & 1338.78  \\
            FastV  & 53.70 & 56.80 & 47.10 & 1275.41 \\
            SparseVLM  & 50.60 & 69.00 & 22.70 & 1289.92 \\
            \mi &    \sota{55.64} & \sota{71.79} & \sota{53.19} & \sota{1378.49} \\
            \hline
            \textit{\textbf{Qwen3VL-30B}} &  62.51 & 95.29 & 83.20 & 1815.29\\
            \rowcolor{lightgray}
            \multicolumn{5}{c}{\textit{keep 50\%}} \\ 
            \rowcolor{white}
            Random & 55.22 & 91.18 & 34.89 & 1755.03  \\
            cosine &  56.47 & 88.65 & 29.96 & 1791.04  \\
            Attention &  56.02 & 88.98 & 30.02 & 1799.54  \\
            \mi &    \sota{56.66}  & \sota{92.27}  & \sota{37.65} & \sota{1822.19}  \\
            \hline
            \end{tabular}
        }
\end{table}

\begin{table}[h]  
        \centering
        \setlength{\tabcolsep}{5pt} 
        \caption{Efficiency comparison on LLaVA1.5-7B.}
        \label{tab: app_efficiency}
        \resizebox{0.49\textwidth}{!}{
            \begin{tabular}{l cc} 
            \hline
            \textbf{Methods} & \textbf{Mem (GB)} & \textbf{Latency (ms)} \\
            \rowcolor{lightgray}
            \hline
            \multicolumn{3}{c}{\textit{LLaVA1.5-7B/keep 32}} \\ 
            \rowcolor{white}
            SparseVLM &   18.12  &  93.33$_{\pm 0.44}$   \\
            VisPruner & 14.35   & 89.98$_{\pm 0.39}$   \\
            DART &  13.94 & 87.56$_{\pm 0.47}$   \\
            CDPruner &  14.61   &  83.22$_{\pm 0.38}$    \\
            \mia & 14.35 & 77.36$_{\pm 0.40}$  \\
            \mi$_{\lambda=1}$ &  \sota{13.90}   &  76.87$_{\pm 0.32}$   \\
            \mi$_{\lambda=1, \mathrm{Flash}}$ &  \sota{13.90}   &  \sota{75.97$_{\pm 0.28}$}   \\
            \mi$_{\lambda=0.5}$ &  13.90   &  78.71$_{\pm 0.34}$    \\
            \mi$_{\lambda=0.5, \mathrm{Flash}}$ &  13.90   &  78.00$_{\pm 0.31}$    \\
            \hline
            \end{tabular}
        }

\end{table}

\begin{table*}[h]
\centering
\caption{Comparison with other MI-based methods (LLaVA1.5-7B, keep 64).}
\setlength{\tabcolsep}{5pt} 
\begin{tabular}{l cccccc} 
\hline
 & \textbf{GQA} &\textbf{VQA$_\mathrm{v2}$} & \textbf{Vizwiz}  & \textbf{MMMU} & \textbf{MMB} & \textbf{Latency (ms)} \\
\hline
TrimTokenator \citep{zhang2025trimtokenator}  &  -  &  -  &  -  &  36.1  &  61.3  &  - \\
AutoPrune \citep{wang2025autoprune} & \subsota{57.7} & 74.8 & \sota{53.6} & 36.0 & \sota{63.6} & 89.6 \\
MMTok \citep{dong2025mmtok} &  \sota{58.3}  &  {75.2}  &  53.1  &  36.1  &  61.2  &  \subsota{77.7} \\
\textbf{MI-Pruner}  &  56.9  &  \subsota{75.2}  &  53.1  &  \sota{36.3}  &  61.9  &  \sota{77.1} \\
\textbf{MI-Attention}   &  57.0  &  \sota{75.5}  &  \subsota{53.4}  &  \subsota{36.1}  &  \subsota{62.1}  &  85.1 \\
\hline
\end{tabular}
\label{tab: results_mi}
\end{table*}

\subsection{Efficiency Analysis}
\label{subsec:app_exp_efficiency}
We evaluate the end-to-end efficiency by tracking GPU memory usage and latency throughout the process, including visual encoding, LLM prefilling, and subsequent decoding.
To ensure stable results, we conduct 10 warm-up runs and report the average GPU memory usage and latency over 100 repetitions. 

Here, we extend Tab.~\ref{tab: efficiency_llava} in the main paper to $N_\mathrm{keep}=32$ in Tab.~\ref{tab: app_efficiency}, and report the results on FlashAttention2 \citep{dao2023flashattention2}. Among diversity-based methods, DART \citep{wen2025stop} and CDPruner \citep{zhang2025cdpruner} are faster than salience-based methods, but less efficient than~\mi. As analyzed in the original paper \citep{zhang2025cdpruner}, the computational bottleneck of CDPruner lies in its DPP MAP inference, which incurs a complexity of $\mathcal{O}(N_VN_\mathrm{keep}^2)$. In comparison, \mi's complexity is $\mathcal{O}(N_V \cdot \max(N_T,N_\mathrm{keep}))$ for full scoring and $\mathcal{O}(N_V  N_T)$ for relevance-based sorting. 
Applying our MI-based scores in VisPruner, \mia~achieves the same GPU memory usage as VisPruner while exhibiting lower latency. Moreover, \mi~speeds up the inference further with FlashAttention cores, showing good compatibility.

\subsection{Spatial Prior}
\label{subsec:app_exp_more}

As acknowledged in the limitations, we adopt the uniform spatial prior assumption for generalization. We claim that such an unbiased prior shows most advantages over many complex alternatives.
To provide an ablation study, we introduce a "central object" prior by a weight factor $\mathbf{w}_{ij}$ for LLaVA1.5-7B, where an image is partitioned into $24\times 24$ tokens. Denote $\mathbf{v}_{ij}$ as an image token in the 2D feature map and $\mathbf{v}_\mathrm{center}$ as the center, we have:
\begin{align}
    \mathbf{w}_{ij} &= \exp\left( -\frac{\|\mathbf{v}_{ij} - \mathbf{v}_{\mathrm{center}}\|^2}{2\sigma^2} \right),
    \\
    p(\mathbf{v}_{ij})&=\mathrm{softmax}(\mathbf{w}_{ij}),
\end{align}
where $\sigma$ controls the decay rate and is measured in tokens. Under the default uniform prior, $p(\mathbf{v}_{ij}){=}\frac{1}{N_V}$.
As shown in Tab.~\ref{tab: results_prior}, the central prior doesn’t improve the overall performance, especially for OCR-related tasks (\textit{e.g.} VQA$_\mathrm{text}$). The rationale behind it is that forcing an "object-centric" spatial prior makes MLLMs over-focus on the center and neglect the fine-grained details scattered across the edges. In comparison, the uniform setting introduces less bias to maintain the best performance. 
A more complicated spatial prior and data-dependent tuning might help in some cases.

\begin{table}[h]
\centering
\caption{Extended ablation on spatial prior (LLaVA1.5-7B, keep 64). $\sigma$ controls the decay rate measured in token-wise distance.}
\setlength{\tabcolsep}{5pt} 
\begin{tabular}{l cccc} 
\hline
 & \textbf{GQA}  & \textbf{SQA} & \textbf{VQA$_\mathrm{v2}$} & \textbf{VQA$_\mathrm{text}$}   \\
\hline
Central$_{\sigma{=}2}$ &  55.80 & 68.02 & 74.86 & 52.11 \\
Central$_{\sigma{=}5}$ & 55.38 & 67.50 & 74.24 & 51.39 \\
\textbf{MI-Pruner}  & \sota{56.88} & \sota{69.81} & \sota{75.22} & \sota{54.91} \\
\hline
\end{tabular}
\label{tab: results_prior}
\end{table}

\begin{figure*}[h]
\begin{subfigure}{0.48\linewidth}
    \centering
    \includegraphics[width=\linewidth]{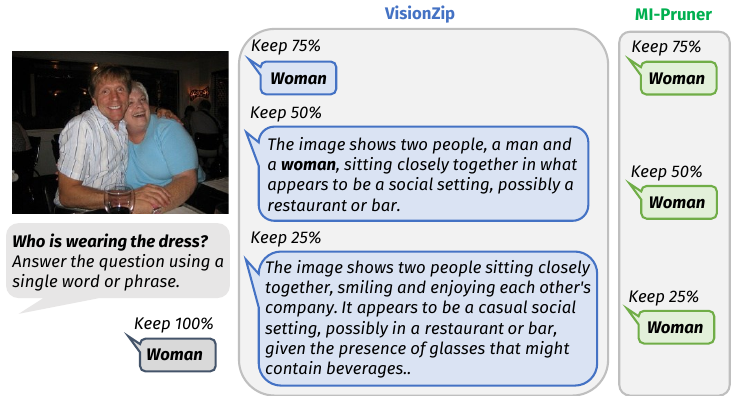}
    \caption{Object}
\end{subfigure}
\begin{subfigure}{0.48\linewidth}
    \centering
    \includegraphics[width=\linewidth]{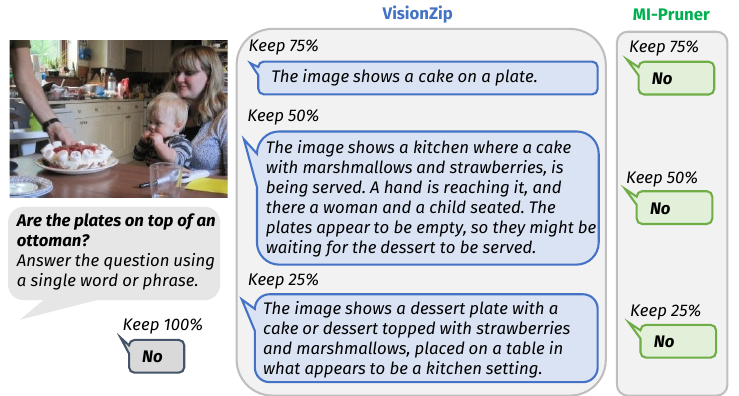}
    \caption{Spatial relationship} 
\end{subfigure}
\caption{Dialogue examples on Qwen2.5VL-7B (GQA).}
\label{fig:qwen2_qa}
\end{figure*}

\subsection{Generalization}
\label{subsec:app_exp_general}
Our method is applicable to any Enc-MLP-Dec architectures, while previous methods rely on specific vision encoders. For instance, VisPruner \citep{zhang2025beyond} necessitates [\texttt{CLS}] as attention measures, therefore, fails to adapt to the latest QwenVL series.
Although VisionZip \citep{yang2025visionzip} provides implementation for Qwen2.5VL, it suffers from a significant decline in instruction-following capacity at high pruning rates, as shown in Fig.~\ref{fig:qwen2_qa}. Due to its limited generalization and lack of theoretical guarantees, the model becomes increasingly prone to hallucinations under aggressive compression, \textit{e.g.}, keeping 25\% vision tokens.

\subsection{POPE Results}
\label{subsec:app_exp_pope}
Tab.~\ref{tab:llava_results}
and \ref{tab:qwen3_results} show the average results of POPE. In the appendix, we report detailed performance under random/popular/adversarial settings, see Tab.~\ref{tab:app_pope_results_llava} and \ref{tab:app_pope_results_qwen3}. The latest Qwen3VL outperforms LLaVA1.5 across three settings, while both fall short of adversarial settings. The results after pruning hold the same tendency. Nevertheless, our method demonstrates remarkable resilience in adversarial settings, outperforming the baselines by a significant margin under different budgets. This success highlights our approach’s acute sensitivity in capturing essential visual features regarding the prompt.

\begin{table*}[h]
\centering
\setlength{\tabcolsep}{5pt}
\caption{\textbf{POPE performance comparison under 3 settings on LLaVA1.5-7B.} We report accuracy (Acc), F1 scores and Yes-rate. }
\begin{tabular}{l ccc ccc ccc ccc}  
\hline
\multirow{2}{*}{\textbf{Methods}} & \multicolumn{3}{c}{\textit{Random}} & \multicolumn{3}{c}{\textit{Popular}} & \multicolumn{3}{c}{\textit{Adversarial}} & \multicolumn{3}{c}{\textbf{\textit{Average}} }\\
& \textbf{Acc} $\uparrow$ & \textbf{F1} $\uparrow$ & \textbf{Yes} & \textbf{Acc} $\uparrow$ & \textbf{F1} $\uparrow$ & \textbf{Yes} & \textbf{Acc} $\uparrow$ & \textbf{F1} $\uparrow$ & \textbf{Yes} & \textbf{Acc} $\uparrow$ & \textbf{F1} $\uparrow$ & \textbf{Yes} \\
\hline
\textit{\textbf{LLaVA1.5-7B}} & 89.60 & 0.90 & 0.51 & 86.20 & 0.87 & 0.55 & 79.73 & 0.82 & 0.61 & 85.18 & 0.86 & 0.56 \\
\rowcolor{lightgray}
\multicolumn{13}{c}{\textit{keep 128}} \\ 
\rowcolor{white}
Random & 85.23 & 0.83 & 0.37 & 84.07 & 0.82 & 0.38 & 80.80 & 0.79 & 0.41 & 83.37 & 0.81 & 0.39 \\
\cosine & 87.43 & 0.86 & 0.42& 86.30 & 0.85 & 0.43 & 82.60 & 0.82 & 0.46 & 85.44 & 0.84 & 0.44 \\
\mia & 88.31 & 0.87 & 0.41 & 86.63 & 0.86 & 0.42 & 84.11 & 0.83 & 0.44 & 86.35 & 0.85 & 0.42 \\
\mi & 88.43 & 0.87 & 0.42 & 86.90 & 0.86 & 0.43 & 84.07 & 0.83 & 0.46 & 86.47 & 0.85 & 0.44 \\
\rowcolor{lightgray}
\multicolumn{13}{c}{\textit{keep 64}} \\ 
\rowcolor{white}
Random &  80.53  & 0.76 & 0.33 &  80.57  & 0.77 & 0.33 &  78.37  & 0.75 & 0.36 & 79.82 & 0.76 & 0.34 \\
\cosine & 85.80 & 0.84 & 0.40 & 85.03 & 0.84 & 0.41 &  81.90 & 0.81 & 0.44 & 84.24 & 0.83 & 0.42 \\ 
\mia  & 86.53 & 0.85 & 0.40 & 85.60 & 0.84 & 0.40  & 83.00 & 0.82 & 0.43 & 85.04 & 0.84 & 0.41  \\
\mi &  86.83  & 0.85 & 0.41 & 85.63 & 0.84 & 0.42 &  82.27  & 0.81 & 0.45 & 84.91 & 0.83 & 0.43 \\
\rowcolor{lightgray}
\multicolumn{13}{c}{\textit{keep 32}} \\ 
\rowcolor{white}
Random & 74.93 & 0.67 & 0.26 & 75.43 & 0.68 & 0.28 & 73.23 & 0.66 & 0.30 & 73.23 & 0.66 & 0.30 \\
\cosine & 83.87 & 0.82 & 0.38& 83.20 & 0.81 & 0.39 & 80.07 & 0.78 & 0.42 & 80.07 & 0.78 & 0.42 \\
\mia & 84.37 & 0.82 & 0.37 & 83.23 & 0.81 & 0.39 & 80.83 & 0.79 & 0.41 & 82.81 & 0.81 & 0.39 \\
\mi & 85.07 & 0.83 & 0.39 & 83.87 & 0.82 & 0.40 & 80.60 & 0.79 & 0.43 & 83.18 & 0.81 & 0.41 \\
\hline
\end{tabular}
\label{tab:app_pope_results_llava}
\end{table*}

\begin{table*}[h]
\centering
\setlength{\tabcolsep}{5pt} 
\caption{\textbf{POPE performance comparison under 3 settings on Qwen3VL.} We report accuracy (Acc), F1 scores and Yes-rate. }
\begin{tabular}{l ccc ccc ccc ccc}  
\hline
\multirow{2}{*}{\textbf{Methods}} & \multicolumn{3}{c}{\textit{Random}} & \multicolumn{3}{c}{\textit{Popular}} & \multicolumn{3}{c}{\textit{Adversarial}} & \multicolumn{3}{c}{\textbf{\textit{Average}} }\\
& \textbf{Acc} $\uparrow$ & \textbf{F1} $\uparrow$ & \textbf{Yes} & \textbf{Acc} $\uparrow$ & \textbf{F1} $\uparrow$ & \textbf{Yes} & \textbf{Acc} $\uparrow$ & \textbf{F1} $\uparrow$ & \textbf{Yes} & \textbf{Acc} $\uparrow$ & \textbf{F1} $\uparrow$ & \textbf{Yes} \\
\hline
\textit{\textbf{Qwen3VL-2B}}  & 92.37 & 0.92 & 0.44 & 89.53 & 0.89 & 0.47  & 87.77 & 0.88 & 0.49  & 89.89 & 0.90 & 0.47  \\
\rowcolor{lightgray}
\multicolumn{13}{c}{\textit{keep 50\%}} \\ 
\rowcolor{white}
Random & 91.67 & 0.91 & 0.44 & 88.47 & 0.88 & 0.46  & 86.13 & 0.86 & 0.48  & 88.76 & 0.88 & 0.46   \\
Attention &  92.33 & 0.92 & 0.44 &  89.40 & 0.89 & 0.47  & 86.90 & 0.87 & 0.50 & 89.54 & 0.89 & 0.47   \\
\cosine & 92.63 & 0.92 & 0.45 &  88.77 & 0.89 & 0.49 & 86.37 & 0.87 & 0.51 & 89.26 & 0.89 & 0.48   \\
\mi &  92.94 & 0.93 & 0.45 &  89.33 & 0.89 & 0.48 & 87.07 & 0.87 & 0.50 & 89.78 & 0.90 & 0.48 \\
\mia &   92.77 & 0.92 & 0.45 & 89.40 & 0.89 & 0.48 & 87.03 & 0.87 & 0.50 & 89.73 & 0.89 & 0.48 \\
\rowcolor{lightgray}
\multicolumn{13}{c}{\textit{keep 25\%}} \\ 
\rowcolor{white}
Random & 88.10 & 0.87 & 0.40 &84.97 & 0.84 & 0.42 &  83.70 & 0.83 & 0.45  & 85.59 & 0.85 & 0.42   \\
Attention & 88.20 & 0.87 & 0.40 &86.43 & 0.85 & 0.41 & 84.53 & 0.83 & 0.43  &  86.39 & 0.85 & 0.41  \\
\cosine & 91.33 & 0.91 & 0.44 &  88.13 & 0.88 & 0.47  &85.37 & 0.85 & 0.49  & 88.28 & 0.88 & 0.47  \\
\mi &   92.73 & 0.92 & 0.45 &89.13 & 0.89 & 0.48 & 86.67 & 0.87 & 0.51  &  89.51 & 0.89 & 0.48\\
\mia & 92.07 & 0.92 & 0.4 & 88.93 & 0.89 & 0.47 & 87.13 & 0.87 & 0.50 & 89.38 & 0.89 & 0.46 \\
\hline
\textit{\textbf{Qwen3VL-8B}}  &  90.77 & 0.90 & 0.42 & 88.67 & 0.88 & 0.44   & 87.07 & 0.87 & 0.46 & 88.84 & 0.88 & 0.44 \\
\rowcolor{lightgray}
\multicolumn{13}{c}{\textit{keep 50\%}} \\ 
\rowcolor{white}
Random & 89.93 & 0.89 & 0.41 & 88.20 & 0.87 & 0.44  & 86.63 & 0.86 & 0.45  & 88.25 & 0.87 & 0.43   \\
Attention &  90.70 & 0.90 & 0.42& 88.73 & 0.88 & 0.44 & 87.13 & 0.87 & 0.46  & 88.85 & 0.88 & 0.44   \\
\cosine &  91.07 & 0.90 & 0.43 & 88.67 & 0.88 & 0.45 & 87.10 & 0.87 & 0.47  & 88.95 & 0.88 & 0.45   \\
\mi &  91.43 & 0.91 & 0.45 & 89.00 & 0.89 & 0.43 & 87.23 & 0.87 & 0.47  &  89.22 & 0.89 & 0.45   \\
\mia &  90.97 & 0.90 & 0.42 & 88.93 & 0.88 & 0.45 & 87.10 & 0.87 & 0.46  & 89.00 & 0.88 & 0.44 \\
\rowcolor{lightgray}
\multicolumn{13}{c}{\textit{keep 25\%}} \\ 
\rowcolor{white}
Random & 87.20 & 0.86 & 0.39   & 85.10 & 0.84 & 0.41  &  84.63 & 0.83 & 0.42  & 85.64 & 0.84 & 0.41  \\
Attention & 89.60 & 0.89 & 0.41  &  87.70 & 0.87 & 0.43 & 85.93 & 0.85 & 0.44 & 87.74 & 0.87 & 0.43   \\
\cosine & 90.43 & 0.90 & 0.43  & 87.40 & 0.87 & 0.46   &  86.10 & 0.86 & 0.47 & 87.98 & 0.88 & 0.45  \\
\mi &  90.70 & 0.90 & 0.43 & 88.77 & 0.88 & 0.45 &  87.00 & 0.87 & 0.46  & 88.82 & 0.88 & 0.45\\
\mia &   90.93 & 0.90 & 0.42 & 88.55 & 0.88 & 0.45 & 87.00 & 0.86 & 0.46 & 88.82 & 0.88 & 0.44 \\

\hline
\end{tabular}
\label{tab:app_pope_results_qwen3}
\end{table*}

\section{Further Discussions}
\label{sec:app_discussion}

\subsection{Interpretation of MI-guided Pruning}
\label{subsec:app_understand}

We study the Mutual Information between an event $\mathbf{v}_i$ and all events $\mathbf{t}_j \in \mathcal{T}$. The normalization notation $\tilde{\cdot}$ is omitted.
This local MI measures the relevance between an image token $\mathbf{v}_i$ and all tokens $\mathbf{t}_j \in \mathcal{T}$ in prompts, which is also formulated as the KL-divergence from the marginal distribution $p(\mathcal{T})$ to the conditional distribution $p(\mathcal{T}|\mathbf{v}_i)$:
\begin{align}
    \mathrm{MI}(\mathcal{T};\mathbf{v}_i) &= \sum_{j=1}^{N_T} p(\mathbf{t}_j|\mathbf{v}_i) \log \frac{p(\mathbf{t}_j|\mathbf{v}_i)}{p(\mathbf{t}_j)}
    \\&= 
    D_\mathrm{KL}(p(\mathcal{T}|\mathbf{v}_i) \| p(\mathcal{T})) 
\end{align}
However, estimating these two probabilities $p(\mathbf{t}_j|\mathbf{v}_i)$ and $p(\mathbf{t}_j)$ is non-trivial in high-dimensional projection spaces. Previous approaches often resort to the matrix rank, determinant or kernel-based diversity measures \citep{zhang2025cdpruner}, which suffer from computational expense for matrix inversion or decomposition, and are highly sensitive to numerical instability. In contrast, we adopt the cosine-based Boltzmann distribution with a softmax operation, maintaining a balance between theoretical soundness and computational tractability.

\subsection{Pruning Stages}
\label{subsec:app_stage}
The data processing inequality \citep{jaynes1957information} indicates "processing cannot increase information". Building on this principle, we perform pruning in the projection space, where visual and textual features are semantically aligned but have not yet undergone crossmodal interaction, thereby minimizing the risk of introducing noisy dependencies. Our motivation is similar to the Q-Former \citep{li2023blip}, while theoretically grounded without extra training.
It's also possible to conduct MI-guided pruning merely on the image features, \textit{i.e.}, after the vision encoder like VisPruner \citep{zhang2025beyond}. However, the prompt-agnostic pruning falls short of a truly multimodal setting, since it overlooks the text information.

\subsection{Pruning Levels}
\label{subsec:app_level}
Model pruning can be applied at different levels, including weights, architecture and tokens (features).
The weight compression tends to be connected with special hardware for acceleration, and the architecture pruning includes layer-level and head-level. Similar to our work, \citet{fan2021layer} leverages Mutual Information to prune layers in a top-to-bottom manner.
Recent work \citep{voita2019analyzing,wang2021spatten,kang2025your} points out that only a few attention heads are necessary in transformer blocks. 

\subsection{Settings and Influences}
\label{subsec:app_future}
Following existing benchmarks \citep{zhang2024sparsevlm,chen2024image}, we test our method under a given token budget (\textit{e.g.} $N_\mathrm{keep}=50\%$ means keeping 50\% visual tokens). However, a more practical and challenging setup would be, \textit{given a minimum performance threshold and a maximum computation limit, the pruning algorithm automatically decides the trade-off between accuracy and efficiency.} We consider this "dynamic budget" setting as our future work, \textit{i.e.}, to determine the optimal number of tokens for each input adaptively. In addition, pruning holds the potential to mitigate hallucination \citep{che2025hallucinatory}. Our experiments on LLaVA1.5 \citep{liu2024improved} demonstrate that heavy pruning on POPE \citep{li2023evaluating} doesn’t degrade performance compared to the full-budget setting, and on Qwen3VL \citep{Qwen3VL}, \mi~even leads to improved POPE performance. We attribute these gains to the reduced reasoning difficulty, where less visual information attends to LLM decoding. However, inappropriate pruning can exacerbate hallucinations, as illustrated in Fig.~\ref{fig:qwen2_qa}. These findings suggest that strategic token pruning represents a promising direction for mitigating hallucination in MLLMs.

\section{Datasets}
\label{sec:app_datasets}

\paragraph{GQA} \citep{hudson2019gqa}. 
The GQA benchmark is designed to evaluate structured spatial understanding and reasoning within visual scenes. Beyond images and questions, it provides comprehensive scene graph annotations, offering structured representations of objects, attributes, and their inter-relationships. For evaluation, we report the accuracy on the test-dev split, which comprises \colorbox{beige}{12,578} image-question pairs.

\paragraph{SQA} \citep{lu2022learn}. 
The ScienceQA benchmark employs multiple-choice questions to assess a model's performance in the scientific domains. The dataset spans three primary subjects—natural science, language science, and social science—and features a hierarchical structure organized by topic, category, and skill. This hierarchy comprises 26 topics, 127 categories, and 379 skills. While the questions are accompanied by relevant illustrations, a portion of the dataset is text-only. For our evaluation, we focus on the SQA-IMG subset, which consists of \colorbox{beige}{2,017} multimodal pairs where both images and questions are present.

\paragraph{VQA$_\mathrm{text}$} \citep{singh2019towards}.
The VQA$_\mathrm{text}$ benchmark works to assess a model’s proficiency in reading and reasoning over visual text. It emphasizes the integration of Optical Character Recognition (OCR) with natural language understanding. The images, primarily sourced from OpenImages-v3 \citep{openimages}, feature diverse real-world scenarios—such as street signs, billboards, and product packaging—that are rich in textual content. Alongside the raw images, ground-truth OCR tokens are provided as auxiliary input. To arrive at the correct answer, models must either extract text directly from the image or perform contextual reasoning based on the identified characters. We report evaluation results on the validation set, which comprises \colorbox{beige}{5,000} image-question pairs.

\paragraph{Seed} \citep{li2024seed}. 
The evaluation of the Seed benchmark covers 12 distinct aspects. It assesses the models’ ability across spatial understanding (\textit{e.g.}, scene understanding and instance identity) and temporal understanding (\textit{e.g.}, action recognition and action prediction). 
The image subset used in this paper includes \colorbox{beige}{14,233} image-question pairs.

\paragraph{MME$_\mathrm{P}$} \citep{fu2025mme}.
The MME benchmark encompasses 14 subtasks, including perception and cognition categories. We focus on the perception part (MME$_\mathrm{P}$), which includes OCR, coarse-grained recognition (presence, count, position and color) and fine-grained recognition (posters, celebrities, scenes, landmarks and artworks).
All of the \colorbox{beige}{2,374} questions belong to binary judgment tasks.

\paragraph{POPE} \citep{li2023evaluating}. 
The POPE benchmark evaluates the object hallucination in large vision-language models with "Yes-or-No" questions. The images are from MSCOCO dataset \citep{lin2014microsoft}, and the questions are about whether
a specific object is present in the image. We report the average Accuracy, F1 score and Yes-rate
across three different sampling strategies in the main paper. Notably, we use the latest version, where three strategies include random (3,000), popular (3,000) and adversarial (3,000), leading to overall \colorbox{beige}{9,000} image-question pairs.


\paragraph{Video datasets.} Video benchmarks extend the 
image-based VQA into the video domain. The GIFs in TGIF-QA \citep{jang2017tgif} are based on Tumblr GIF dataset \citep{li2016tgif}, while MSVD-QA \citep{xu2017video} and MSRVTT-QA \citep{xu2017video} incorporate Microsoft Research Video Description Corpus \citep{chen2011collecting} and Microsoft Research Video to Text \citep{xu2016msr} dataset, respectively.
Following previous work \citep{yang2025visionzip}, we test on the first \colorbox{beige}{1,000} samples of all three video-QA datasets. All of them are scored by GPT-3.5-Turbo \citep{openai2023gpt35turbo}.